\documentclass[sigconf,screen]{acmart}

\usepackage[utf8]{inputenc}
\usepackage[linesnumbered,ruled,vlined]{algorithm2e}
\usepackage{algpseudocode}
\usepackage{subfig}
\usepackage{seqsplit}
\usepackage{multirow}
\usepackage{listings}
\usepackage{placeins}
\usepackage{enumitem}
\usepackage{etoolbox}

\setlist[itemize]{leftmargin=2em}

\AtBeginDocument{%
  }

\setcopyright{acmcopyright}
\copyrightyear{2024}
\acmYear{2024}
\acmDOI{XXXXXXX.XXXXXXX}
\acmConference[ISSTA 2024]{ACM SIGSOFT International Symposium on Software Testing and Analysis}{16-20 September, 2024}{Vienna, Austria}

\acmPrice{15.00}
\acmISBN{978-1-4503-XXXX-X/23/08}

\SetAlgoSkip{}

\begin{document}

\title{Dance of the ADS: Orchestrating Failures through Historically-Informed Scenario Fuzzing}


\author {Tong Wang}
\orcid{0000-0001-6981-916X}
\affiliation{%
	\institution{Academy of Military Sciences}
	\city{Beijing}
	\country{China}
}

\author {Taotao Gu}
\affiliation{%
	\institution{Academy of Military Sciences}
	\city{Beijing}
	\country{China}
}

\author {Huan Deng}
\affiliation{%
	\institution{Academy of Military Sciences}
	\city{Beijing}
	\country{China}
}

\author {Hu Li}
\affiliation{%
	\institution{Academy of Military Sciences}
	\city{Beijing}
	\country{China}
}

\author {Xiaohui Kuang}
\affiliation{%
	\institution{Academy of Military Sciences}
	\city{Beijing}
	\country{China}
}

\author {Gang Zhao}
\authornote{\footnotesize{Corresponding author (zemell@foxmail.com)}\vspace{-5pt}}
\affiliation{%
	\institution{Academy of Military Sciences}
	\city{Beijing}
	\country{China}
}


\begin{abstract}
	As autonomous driving systems (ADS) advance towards higher levels of autonomy, orchestrating their safety verification becomes increasingly intricate. This paper unveils \textit{ScenarioFuzz}, a pioneering scenario-based fuzz testing methodology. Designed like a choreographer who understands the past performances, it uncovers vulnerabilities in ADS without the crutch of predefined scenarios. Leveraging map road networks, such as OPENDRIVE, we extract essential data to form a foundational scenario seed corpus. This corpus, enriched with pertinent information, provides the necessary boundaries for fuzz testing in the absence of starting scenarios. Our approach integrates specialized mutators and mutation techniques, combined with a graph neural network model, to predict and filter out high-risk scenario seeds, optimizing the fuzzing process using historical test data. Compared to other methods, our approach reduces the time cost by an average of 60.3\%, while the number of error scenarios discovered per unit of time increases by 103\%. Furthermore, we propose a self-supervised collision trajectory clustering method, which aids in identifying and summarizing 54 high-risk scenario categories prone to inducing ADS faults. Our experiments have successfully uncovered 58 bugs across six tested systems, emphasizing the critical safety concerns of ADS. 
\end{abstract}



\maketitle
\section{Introduction}
\label{section1}

Rapid advancements in autonomous driving systems are driving the shift from Level 1 to Level 5 autonomy, closely tied with integrating AI technologies like deep neural networks. As ADS evolves, the complexity of their models, the multiplicity of modules, the probabilistic nature of deep learning, extensive parameter counts, and training data volume collectively heighten safety risks and pose challenges in ADS verification and testing. The variety of external environments further complicates these challenges. The National Highway Traffic Safety Administration (NHTSA) reports numerous ADS-related accidents annually, highlighting the urgent need for rigorous safety verification in industry and academia \cite{stafford2022nhtsa}.

Given the complexity of ADS, verifying the safety of individual modules and their interactions is crucial. Scenario verification offers a practical solution, validating the overall system and module interactions. However, constructing diverse real-world scenarios is impractical and costly, leading to the use of simulation engines for customizable scenario creation. While many manufacturers focus on developing scenario libraries for testing, identifying and constructing scenarios that effectively induce ADS errors remains a pressing research issue \cite{tangSurveyAutomatedDriving2022,zhongSurveyScenarioBasedTesting2021,dingSurveySafetyCriticalDriving2022}.

In this context, we draw an analogy between the ADS and a lead actor in a play, aiming to create a "script" that tests the ADS under various challenging conditions and exposes system vulnerabilities.

\textbf{Script}: The key challenge is the script's availability, particularly scenario sources. These originate from traffic videos, accident data, and mandated scenarios, with some studies converting these into simulation scenarios \cite{zhang2023building,gambi2022generating,wangADEPTTestingPlatform2022,xuSafeBenchBenchmarkingPlatform2022}. Others use scenario description languages to translate scenarios into simulations \cite{ding2023causalaf,horelUsingFormalConformance2022,karunakaran2022critical,liSML4ADSOpenDSML2022,norden2020testing,ramakrishnaANTICARLAAdversarialTesting2022,ramakrishnaRiskAwareSceneSampling2022,tahirIntersectionFocusedSituation2022,9251068,zhong2022neural, BehAVExplor_ISSTA_2023}. However, organizing and acquiring these sources is labor-intensive and often limited in scope.

\textbf{Stage Scope}: The range of scenario layouts is another critical aspect. An expansive scope incurs long execution times and computational costs. Randomly selecting points in a city map to find ADS flaws is not only computationally intensive but also adds uncertainty to testing.

\textbf{Stage Elements}: Key elements like roads, signs, and natural features are often fixed in reality, but modifying these in simulations deviates from scenario generation goals and is technically challenging.

\textbf{Stage Lighting}: Lighting, representing weather conditions, is crucial. Simulation platforms like CARLA \cite{Dosovitskiy17} and LGSVL \cite{rong2020lgsvl} effectively emulate diverse weather conditions, essential for testing ADS's visual perception.

\textbf{Supporting Cast}: Vehicles and pedestrians are often catalysts for ADS errors. Their variety and interactions with ADS can lead to accidents and losses.

Some studies focus narrowly on specific components, like road geometries, potentially overlooking comprehensive ADS evaluation or simulation fidelity \cite{gambi2019automatically,zohdinasab2023efficient}. Others concentrate on the movement of supporting actors, using various methods to create adversarial trajectories \cite{9251068,hanselmann2022king,wei2023controllable,von2023deepmaneuver,zhang2022adversarial,zhang2023testing,291108}.

However, some approaches aim to encompass entire scenarios, using formal constraints, parameter search, fuzz testing, and causal generation \cite{horelUsingFormalConformance2022,haq2022efficient,karunakaran2022critical,ramakrishnaANTICARLAAdversarialTesting2022,ramakrishnaRiskAwareSceneSampling2022,norden2020testing,tahirIntersectionFocusedSituation2022}. Such methods often require a predefined scenario base, limiting the scope of tests and necessitating a dedicated scenario repository.

Our approach aims to automatically generate and discover error scenarios in ADS without relying on a predefined script. This broadens the scope and targets the testing process more effectively. Drivefuzz \cite{Kim22DriveFuzz} creates test scenarios by randomly selecting two locations within the map to serve as the starting and ending points for the test vehicle, which can result in excessively large scenario ranges. Additionally, the placement and paths of other vehicles and pedestrians are set without reference information, being randomly placed within a specified distance from the test vehicle's starting point. Such an approach may lead to the generation of scenarios that are not grounded in realistic driving conditions, due to the lack of a suitable, adequate information source for scenario creation.

We focus on leveraging existing simulation platforms and assets to uncover ADS bugs efficiently without predefined scenarios. Our method involves constructing a corpus enriched with relevant information, including specific scenario ranges and compliant paths. Using the Map Crawling Technique, we construct topological maps and extract various data to serve as a scenario seed library. Integrated with CARLA, we design scenario mutators considering the fundamental elements of a scenario, introducing an efficient two-stage mutation strategy and a distance-based proximity sampling method. Coupled with historical data, we convert test scenarios into graph-type data, feeding a Graph Neural Network-based (GNN) evaluation model for seed filtering. This approach ensures model generalizability across different testing systems, combining dynamic and static scenario information.

Our paper makes the following contributions:

\begin{itemize}
	\item We introduce \textit{ScenarioFuzz}, a novel fuzz testing approach for autonomous driving systems that begins without a preset scenario. It uses unique mutators and mutation techniques for comprehensive scenario coverage and is integrated into the CARLA simulator \cite{Dosovitskiy17}, enhancing compatibility with CARLA Leaderboard interfaces \cite{carla_leaderboard}.
	\item We elucidate a method to extract data from map road networks, like OPENDRIVE, which serves as an essential scenario seed corpus. This corpus acts as an invaluable information repository and establishes boundaries for fuzz testing, eliminating the need for predefined starting scenarios.
	\item we employ a GNN model to predict and sieve out high-risk scenario seeds, thus optimizing fuzzing using prior test data.
	\item We propose a self-supervised collision trajectory clustering method. Through this, we identify and categorize 54 high-risk scenarios likely to induce collisions in ADS, offering critical insights for testing and training. Our experiments revealed 58 bugs in six systems, underscoring significant safety concerns.
\end{itemize}

\section{Background}
\label{section2}

\subsection{Autonomous Driving Systems (ADS)}
\label{section2:1}

\begin{figure}[htbp]
	\vspace{-10pt}
	\begin{center}
		\includegraphics[width=0.48\textwidth]{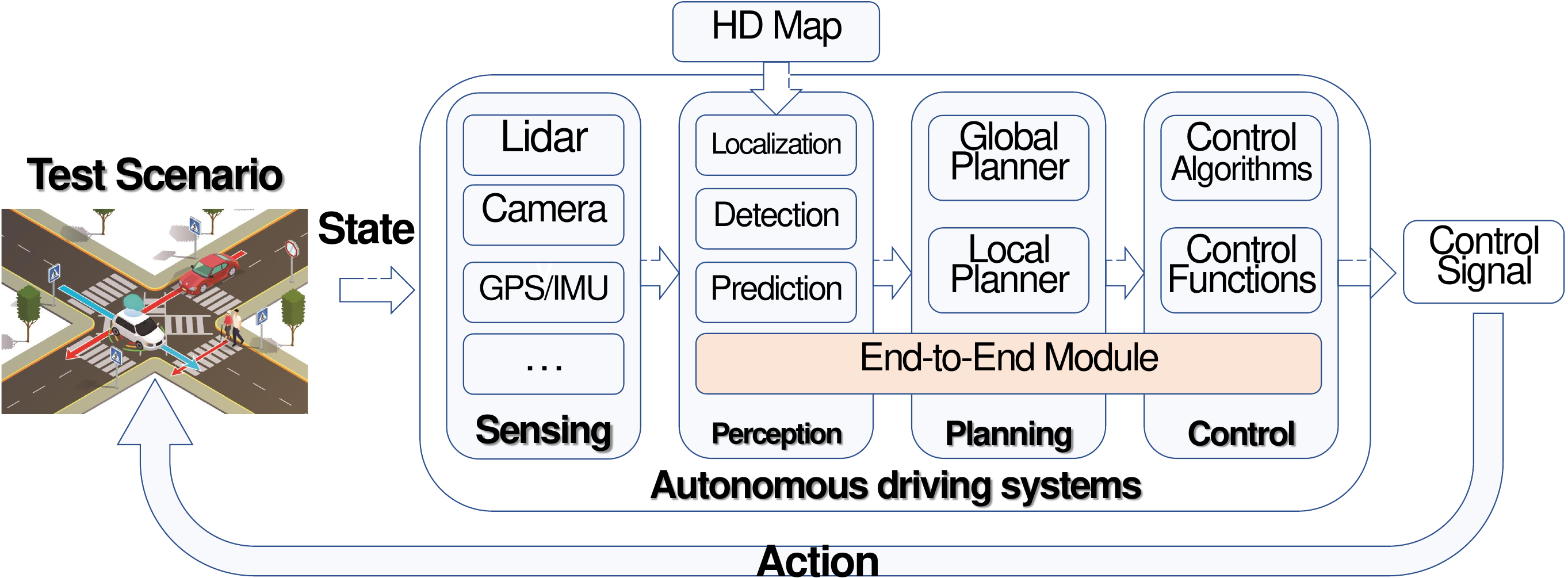}
		\vspace{-10pt}
		\caption{Autonomous driving system and testing scenarios.}
		\label{fig1}
		\vspace{-18pt}
	\end{center}
\end{figure}

Figure \ref{fig1} illustrates the architecture of contemporary ADS, comprised of four main modules: sensing, perception, planning, and control. An alternate version, end-to-end autonomous driving systems, combines the last three modules as a whole. 

The sensing module encompasses various sensors, such as radar, cameras, GPS, and IMU, to collect environmental and vehicle data. Perception involves processing sensor data for vehicle localization, object detection, and behavior prediction. Planning formulates driving routes based on perception results and driving tasks, involving both global and local planners. The control module generates vehicle control signals using various algorithms and functions. While deep learning techniques find use in perception and end-to-end modules, other modules typically employ logic-based or corresponding computational methodologies.

\subsection{Scenarios in the Context of ADS}
\label{section2:2}

In ADS, scenario and scene are two concepts. As defined by S.Ulbrich et al.\cite{ulbrich2015defining}, a scenario is a temporal sequence of scenes, each representing a snapshot of the environment, including entities and their relationships. As shown on the left of Figure \ref{fig1}, the continuous process displayed in this image portrays a typical scenario. The tested vehicle (Ego-car) is driving straight through an intersection, with a potential for collisions with other vehicles moving in the perpendicular direction and pedestrians crossing the road. Simultaneously, traffic lights, speed limit signs, and other traffic infrastructure are present, and the current weather is sunny. A scene, in contrast, is the instantaneous state depicted in the image.

\begin{figure}[htbp]
	\vspace{-10pt}
	\begin{center}
		\includegraphics[width=0.48\textwidth]{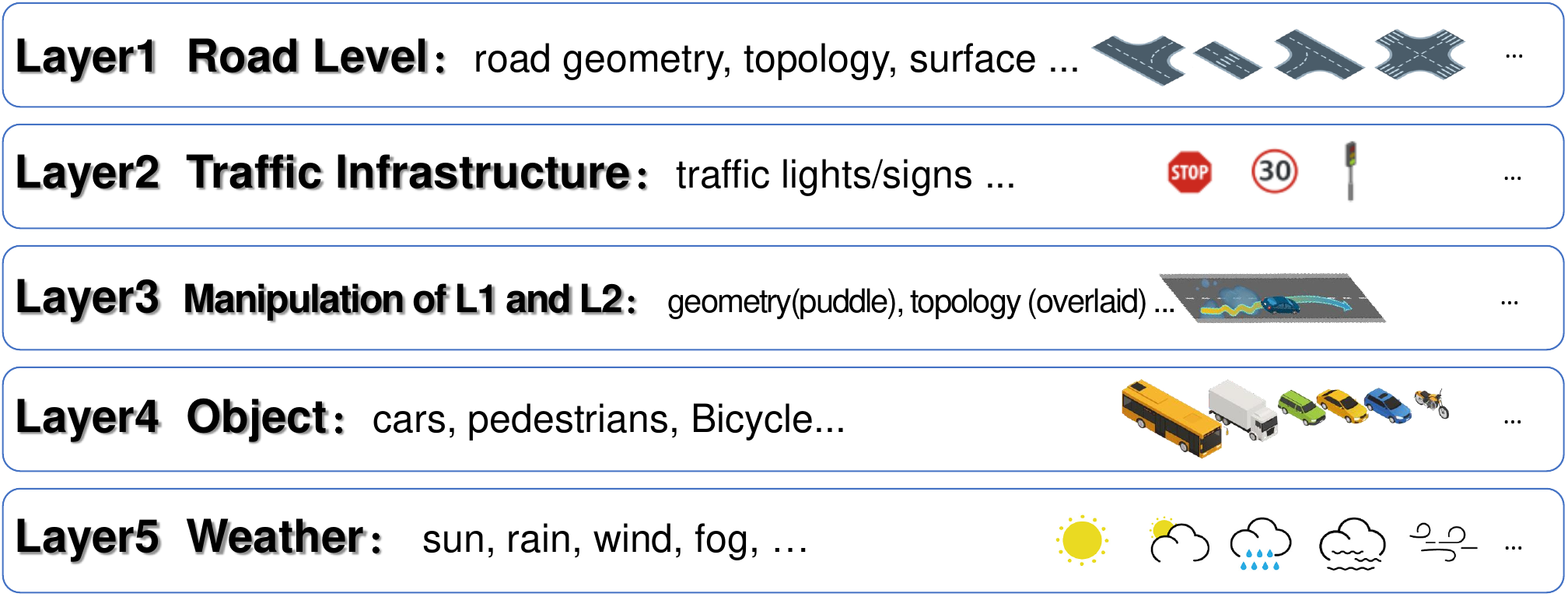}
		\vspace{-10pt}
		\caption{Overview of the scenario layer model \cite{zhongSurveyScenarioBasedTesting2021}.}
		\label{fig2}
		\vspace{-18pt}
	\end{center}
\end{figure}

Bagschik et al. \cite{bagschik2018ontology} proposed a scenario layer model in Figure \ref{fig2}. Layer 1 comprises the road layer, which includes road materials and types. Layer 2 constitutes the traffic infrastructure, encompassing traffic lights, signs, etc. Layer 3 involves operations on Layers 1 and 2, such as puddles on the road. Layer 4 consists of objects in the scenario, such as pedestrians and vehicles, each with a series of attributes like motion state, appearance type, and path. Layer 5 represents weather, including sunlight conditions (intensity and angle), rain conditions (precipitation intensity), and other factors.

In our work using the CARLA simulator, we cannot adjust the Layer 1 setting, but we can identify and extract viable waypoints and paths for various road types and traffic infrastructure, using the map crawling technique (see \S \ref{section4:2}). This process does not alter the road shapes or infrastructure but allows us to obtain detailed information for scenario construction. Layer 3 can include puddles, while Layer 4 allows 26 types of vehicles and pedestrians, color modifications, motion states, and realistic paths. Layer 5 controls eight weather parameters to create diverse conditions. Our method encompasses all levels of scenario composition, ensuring richness and comprehensiveness.

\section{Approach}
\label{section4}

\subsection{Overview of the ScenarioFuzz Method}
\label{section4:1}

\begin{figure*}[htp]
	\begin{center}
		\includegraphics[width=0.95\textwidth]{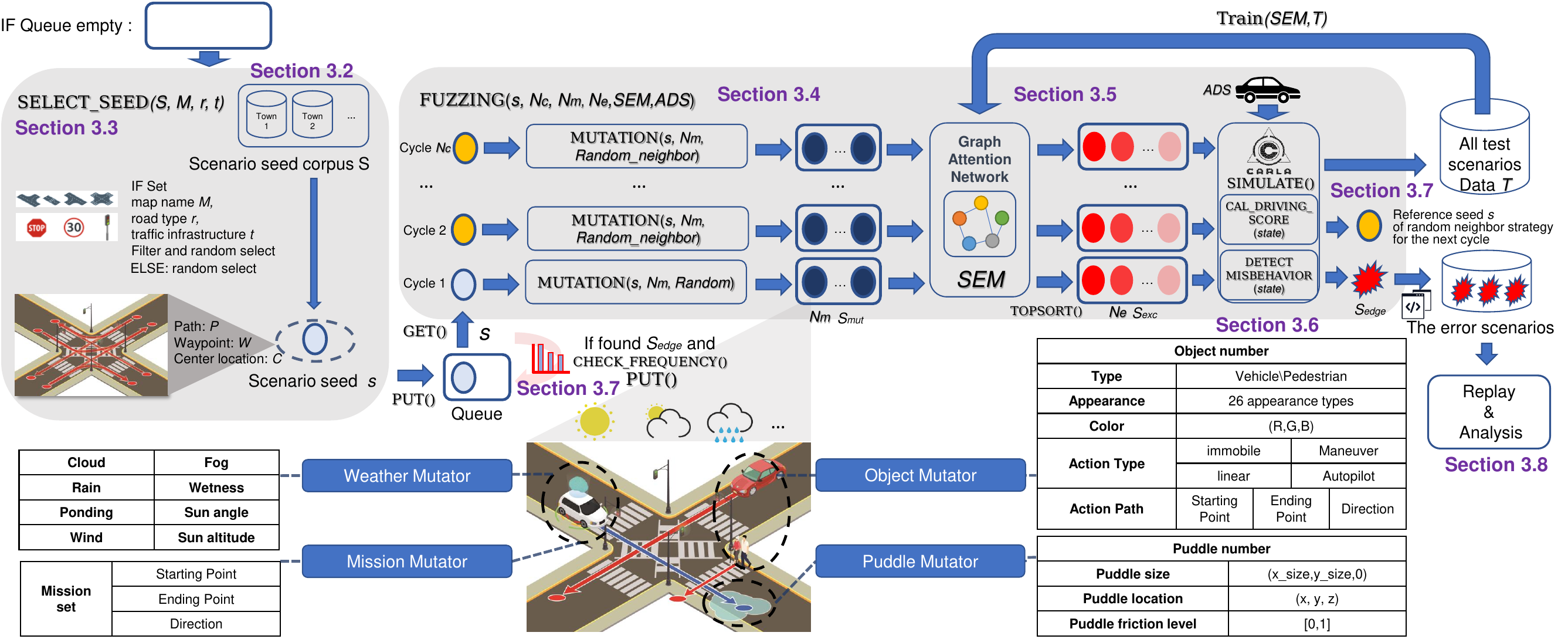}
		\vspace{-10pt}
		\caption{Overview of the architecture and workflow of ScenarioFuzz.}
		\label{fig3}
		\vspace{-10pt}
	\end{center}
\end{figure*}

Our approach leverages map crawling technology to generate an initial scenario seed corpus for fuzzing. The selected seeds then undergo a two-stage mutation strategy using a variety of mutators for the mutations. A scenario evaluation model filters these mutated seeds and selects the one projected to pose the highest risk as the test case. The test case is utilized to build the simulator scenario and integrate it into the ADS. (See Figure \ref{fig3} and Algorithm \ref{alg2}).

In this workflow, the function {\small \textsc{select\_seed}()} filters seeds based on the map name, road type, and traffic infrastructure, and randomly selects a seed from this set. Over $N_c$ mutation cycles, $N_m$ mutated seeds are generated in each cycle using the two-stage {\small\textsc{Mutation}()} process. From the second round, informed random neighbor mutations are conducted. The scenario is mutated by several mutators, including weather, object, puddle, and ADS mission.

Subsequently, the scenario evaluation model ($SEM$, detailed in \S \ref{section4:5}) filters $N_e$ seeds most likely to cause ADS error. This model also undergoes real-time training {\small \textsc{train}()} once the test data $T$ (all systems under test) reaches a certain size. These seeds are used to construct the scenario within the simulator and connect to the ego car. The {\small\textsc{detect\_misbehavior}()} function monitors ADS behavior in real-time and identifies error scenarios. Once an error scenario is identified, the fuzzing operation is stopped and the case is retained. The selection frequency of the corresponding scenario seed is evaluated using the {\small{\textsc{check\_frequency}}()} function, with a roulette-based approach determining the seed's reinsertion into the queue. If a seed stops yielding error scenarios, an alternative seed is chosen for the following rounds. Seeds with high selection frequencies are excluded to prevent local optima and ensure diversity.

{
	
	\scriptsize
	\begin{algorithm}
		
		\setlength\algomargin{8em}
		\caption{ScenarioFuzz}
		\label{alg2}
		\KwIn{Scenario seed corpus $S$, Max cycles $N_c$, Mutation seed nums $N_m$, Max execute seed nums $N_e$, town map name $M$, road type $r$, traffic infrastructure $t$, Scenario Evaluation Model $\textsc{SEM}$, Tested ADS $ADS$, Train size $tr$, Error scenario library $S_{error}$, Test data $T$}
		\KwOut{The error scenarios $S_{error}$}
		\SetKwFunction{Fuzzing}{fuzzing}
		initialize empty queue $q$\;
		$S_{sel} \leftarrow$ \textsc{select\_seed}($S,M,r,t$)\;
		\If{$q$ is empty}{
			$s \leftarrow$ \textsc{random}($S_{sel}$)\;
			\textsc{put}($q,s$)\;
		}
		\While{not at the end of $q$}{
			$s \leftarrow$ \textsc{get}($q$)\;
			$s_{error} \leftarrow$ \Fuzzing($s,N_c,N_m,N_e,SEM,ADS$)\;
			\If{$s_{error}$ is not empty}{
				$S_{error} \gets S_{error} \cup \{s_{error}\}$\;
				\If {\textsc{check\_frequency}($s$)}{
					\textsc{put}($q,s$);}               
			} 
			\If{length($T$) mod $tr$ == 0}{
				\textsc{train}(SEM,$T$)\; 
			}                  		
		}
		
		\SetKwProg{Fn}{Function}{:}{}
		\Fn{\Fuzzing{$s$, $N_c$, $N_m$, $N_e$, \textsc{SEM}, $ADS$}}{
			\For{$i \leftarrow 1$ \KwTo $N_c$}{
				\If{$i=1$}{$S_{mut} \leftarrow$ \textsc{mutation}($s,N_m,random$)\;}
				\Else{$S_{mut} \leftarrow$ \textsc{mutation}($s,N_m,random\_neighbor$)\;}
				\ForEach{$y_{pred},y_{score},s_{pred}$ in \textsc{SEM}($S_{mut}$)}{
					\If{$y_{pred}$ is True}{
						$Y_{pred} \leftarrow (y_{pred},y_{score},s_{pred})$\;
					}
				}
				\textsc{topsort}($Y_{pred}$)\;
				$score \leftarrow 0$\;
				\For{$j \leftarrow 1$ \KwTo $N_e$}{
					$s_{exc} \leftarrow Y_{pred}[j]$\;
					$state \leftarrow$ \textsc{simulate}($s_{exc},ADS$)\;
					\If{ \textsc{detect\_misbehavior}($state$) is True}{
						$T \gets T \cup \{(s_{exc},True)\}$\;
						\Return{$s_{exc}$}\;}
					\Else{
						$T \gets T \cup \{(s_{exc},False)\}$\;
						$score_j \leftarrow$\textsc{cal\_driving\_score}($state$)\;
						\If{$score_j \leq score$}{
							$score \leftarrow score_j$\;
							$s_{error} \leftarrow s_{exc}$}					
						
					}
				}
			}
			\Return{$\emptyset$}
		}
		
	\end{algorithm}
}

\subsection{Building Corpus with Map Crawling}
\label{section4:2}


The corpus serves as the cornerstone of our scenario construction process. Utilizing the map crawling method with the map road networks as input, we construct topological graphs, and then segment and classify roads by their types based on geographical location and topological structure. Furthermore, we extract key data from these segments, thereby constructing an essential scenario seed corpus for our study.

Firstly, we read the OPENDRIVE road networks file for the city map in the CARLA simulator (e.g., Town03 map). Using waypoint connectivity relationships, we construct a topological graph. In CARLA, the API call {\footnotesize \texttt{carla.map.get\_topology(self)}} assembles the gra\-ph. On other platforms, the OPENDRIVE format data can be used to generate the graph by extracting waypoint information and trav\-ersing the{\footnotesize \texttt{<predecessor>}} and {\footnotesize \texttt{<successor>}} elements.

Next, using Algorithm \ref{alg1}, we apply {\small \textsc{get\_traffic\_light}()} to read the traffic light locations in map $M$ and use {\small\textsc{cluster}()} to cluster them. This allows us to obtain scenario regions for intersections with traffic lights. The {\small\textsc{get\_near\_nodes}()} function identifies the nod\-es (i.e., waypoints) closest to the traffic lights, forming the scenario seeds. For unused nodes in the topological graph $G$, we employ {\small$\textsc{cluster}()$} to cluster the nodes by location, obtaining seeds without traffic lights. This prevents misclustering, as pre-selecting intersection scenarios based on traffic light distribution.

The current scenario seeds include the traffic light location $t$ and waypoint information $n$. We use {\small\textsc{classified\_road\_type}()} to read the connectivity of $n$ in $G$, determine the node with the highest connectivity, and ascertain the road type $r$ for the scenario. The rules for this are provided in the supplementary material. The {\small\textsc{get\_path}()} function retrieves feasible motion paths $p$, while {\small\textsc{cal\_center}()} calculates the center position $c$ of all nodes in $n$. {\small \textsc{search\_map}()} collects map information $o$ (e.g., nearby traffic signs and lane change data). Lastly, we input ${r, p, c, o}$ into the scenario seeds, storing them in the seed corpus for each map. The seed corpora are stored in the scenario seed corpus by the map's name. Construction took 223.2s across 5 maps, with time varying based on map size.

\vspace{-10pt}
{	\scriptsize
	\begin{algorithm}
	\setlength\algomargin{8em}
	\caption{Scenario Crawling Algorithm}
	\label{alg1}
	\KwIn{Topology graph $G$, Map information $M$}
	\KwOut{Map Scenario seed corpus $S$}
	initialize an empty corpus $S$\;
	$T \leftarrow$ \textsc{get\_traffic\_light}($M$)\;
	\ForEach{$t$ in \textsc{cluster}($T$)}{
		$n \leftarrow \textsc{get\_near\_nodes}(G, t)$\;
		$S \leftarrow s=\{n,t\}$\;
	}
	$N \leftarrow$ unused nodes in $G$\;
	\ForEach{$n$ in \textsc{cluster}($N$)}{
		$S \leftarrow s=\{n,\emptyset\}$\;
	}
	\ForEach {$s$ in $S$}{
		road\_type $r \leftarrow \textsc{classified\_road\_type}(G,n\in s)$\;
		paths $p \leftarrow \textsc{get\_path}(G,n\in s)$\;
		center location $c \leftarrow \textsc{cal\_center}(n\in s)$\;
		others $o \leftarrow \textsc{search\_map}(M,n\in s)$\;
		$s \leftarrow s \cup \{r,p,c,o\}$\;
	}
	\Return{$S$}
\end{algorithm}
}
\vspace{-15pt}

\subsection{Scenario Seed Selection}
\label{section4:3}

Initial seeds can be further selected from the corpus based on testing requirements. This selection is based on various data, such as city names, road types, and proximity to specific traffic signs. When only a subset of this information is available, the selection is filtered accordingly. The future mutations of these seeds are then guided by the waypoint positions $n$, the paths $p$, and the central location $c$ within the seed scenario.

\subsection{Scenario Seed Mutation}
\label{section4:4}

The mutation process for a scenario seed consists of two essential components: mutators and mutation strategy. Various types of mutators perform mutation operations on the corresponding elements of the scenario, and the configuration of the associated attribute values is dictated by the mutation strategy.

\subsubsection{Mutators}
\label{section4:4.1}

In the context of scenario mutation, we use several mutator types to adjust different scenario aspects. These alterations aim to examine the ADS's resilience and adaptability under various challenging circumstances.

\textbf{Mission Mutator.} This mutator modifies the ego car's mission path. It first selects a location from the waypoint information $n$ in the scenario seed $s$ as the starting point for the test vehicle. It then randomly selects a path from $p$, ensuring the chosen path includes specified starting and ending points based on the initial location. If a specific driving direction, such as left, straight, or right, is required, $p$ is filtered accordingly. 

\textbf{Puddle Mutator.} This mutator generates puddles within the scenario. It begins by establishing the number of puddles. For each puddle, it generates the corresponding size, location, and friction coefficient. The puddles' presence tests the robustness of the ADS control module since irregular ground friction can significantly influence the control signals' effect on the vehicle state.

\textbf{Object Mutator.} This module generates objects within the scenario, spanning pedestrians to various types of vehicles. It first determines the number of objects to be generated within the scenario. Each object is associated with a specific type, i.e., pedestrian or vehicle. In terms of appearance, vehicle objects can represent buses, passenger cars, police cars, and other common models. Similarly, pedestrian objects can depict individuals with diverse clothing and skin tones, with a total of 26 types available. Certain special models, like bicycle or motorcycle riders, are also included in the vehicle category. For some vehicle models, exterior colors can be specified.

Object action types include:

\begin{itemize}
    \item \textbf{Immobile:} Static objects remain at their starting positions.
    \item \textbf{Linear:} Objects follow pre-planned paths, moving at a consistent speed\footnote{Speeds are within [1,4] m/s for pedestrians and [3,10] m/s for vehicles.} set during their generation.
	\item \textbf{Maneuver:} The pre-planned path is segmented, with the number of segments being the integer part of the total path length divided by eight. Each segment encompasses straight and turning types: a deviation exceeding 5 degrees indicates a left turn, while less than -5 degrees signals a right turn; others are considered straight types. Variations are applied to each segment, altering turn angles\footnote{Turn angles range from [current angle -20, current angle +20].} and straight driving speeds\footnotemark[1] accordingly.
    \item \textbf{Autopilot:} Objects dynamically navigate using CARLA's autopilot, complying with traffic regulations.
\end{itemize}

The objects' presence tests the ADS's sensing, perception, and planning modules. It requires successful object detection from the sensing module, accurate object recognition from the perception module, and strategic path planning based on the object's location and expected movement. Any collision between the test vehicle and an object is considered a severe safety violation. Additionally, dynamic interaction between objects could potentially lead to collisions, reflecting unexpected real-world traffic accidents and testing the system's emergency response capabilities.

\textbf{Weather Mutator.} This mutator adjusts eight weather parameters: cloud coverage, rain, ponding, wind, fog, wetness, sun angle, and sun altitude. These parameters can simulate a variety of weather conditions, significantly affecting the ADS's sensing and perception modules. Severe weather conditions can interfere with the usability of signals collected by cameras and lidar and the environmental perception accuracy of visual models.


\subsubsection{Mutation Strategy}
\label{section4:4.2}
Our mutators require mutating multiple attribute parameters, which are further subdivided into discrete and continuous categories. Discrete attributes typically cover integers or specific choices, such as quantity or type. In contrast, continuous attributes typically represent decimal values and necessitate a defined range and precision to generate corresponding values. The mutation strategy guides the selection of these attribute values. Our method utilizes a two-stage mutation strategy.

\textbf{Random:} This strategy randomly selects a value from a predetermined range to serve as the corresponding attribute value. This approach helps to ensure a broad coverage of the input space and potentially discover more anomalies.

\textbf{Random Neighbor:} This strategy generates a new attribute value randomly within a range that extends five steps above and below the current value of each mutator attribute. For discrete values, one 'step' equates to one choice from the set of possible values. For continuous values, one 'step' corresponds to a specific precision, defined by the attribute's measurement scale. After each mutation cycle, the scenario seed with the lowest driving score is chosen. A random neighbor sampling technique based on this seed is then used to draw new inputs from the neighborhood of the existing reference mutation attribute values, thus creating new test cases. This strategy may facilitate a more intensive exploration of the local region of the input space, thereby identifying potential issues.

The incorporation of these two sampling methods allows the implementation of diverse mutation strategies, striking a balance between extensive coverage and localized exploration as required.


\subsection{Mutated Seeds Filtering}
\label{section4:5}

\begin{figure*}[htp]
	\centering
	\includegraphics[width=0.85\textwidth]{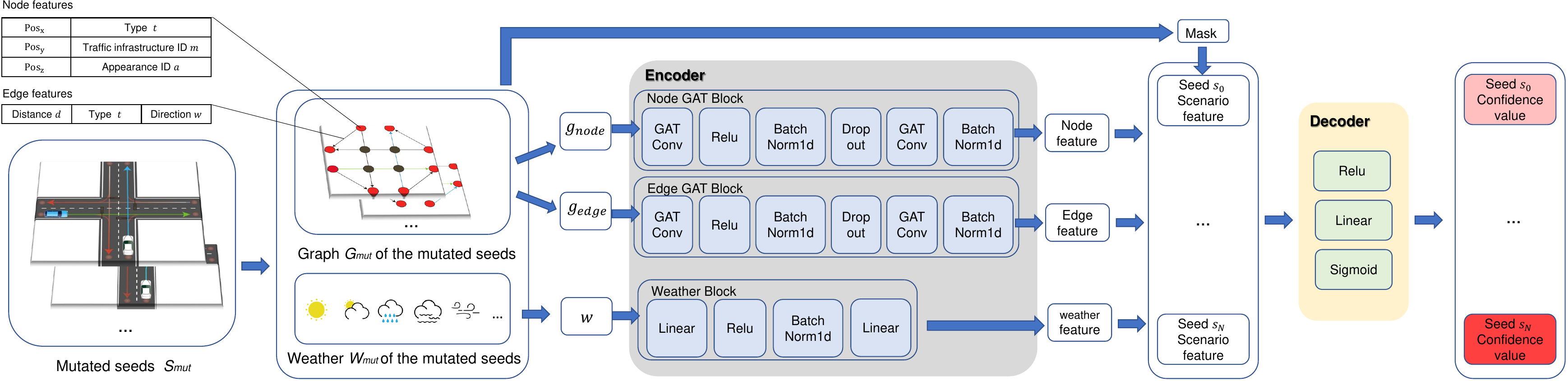}
	\vspace{-10pt}
	\caption {The comprehensive process and architectural diagram of the scenario evaluation model.}
	\label{fig4}
	\vspace{-10pt}
\end{figure*}

Fuzzing often produces numerous mutated seeds, with many not triggering ADS incidents. Each seed requires substantial resources for scenario construction in the simulator and imposes time and computational overheads for testing. With the average execution time per scenario being approximately several minutes, generating error scenarios amidst the wide range of scenario types and parameter space after numerous mutations becomes both costly and challenging. To counter this, we introduce the scenario evaluation model that estimates the likelihood of each mutated scenario seed evolving into an error scenario and assigns a confidence level. This confidence level is used to filter the mutated seeds, selecting those with the highest probability of becoming error scenarios, which helps save on overhead. Furthermore, this model utilizes historical test data for training, effectively repurposing previously collected data, as illustrated on the right side of Figure \ref{fig3}.

Initially, mutated scenario seeds are converted into graph data, as depicted in Figure \ref{fig4}. Each node $n_i$ encapsulates the waypoint information $n$ from the seed (if a traffic light is present, its position information is also stored in the graph data as a node). Each node's features include the relative position of the node with the ego vehicle initial position $p_{n_{i}}=(x_{n_i}-x_{ego},y_{n_i}-y_{ego},z_{n_i}-z_{ego})$, the node type $t_{n_{i}}$, the traffic sign type $m_{n_{i}}$, and the object appearance type $a_{n_{i}}$ located at that point. The last three types are encoded as integers. The node type encompasses the default case, traffic light, start and end points of the ego vehicle, other vehicles and pedestrians. The traffic sign type and the object appearance type at the point are assigned integer values from corresponding ID sets. If none exist, the value corresponding to the default case is used.

The edges between nodes in the graph are established using the path information $p$ from the seed. Each edge $e_{ij}$ represents a path from node $n_{i}$ to $n_{j}$, featuring distance $d_{e_{ij}}$, type $t_{e_{ij}}$, and direction $w_{e_{ij}}$. The distance $d_{e_{ij}}$ is calculated as the 2-norm distance between the nodes. Both the latter two features are integer-encoded. The edge type $t_{e_{ij}}$ includes the default case, the path of the ego vehicle's task, the driving paths of other vehicles, and the movement paths of pedestrians. The direction $w_{e_{ij}}$ encompasses left turn, right turn, straight, and an unknown case for scenarios where accurately classifying the direction of longer paths is challenging.

Corresponding scenario weather parameters $w$ are incorporated into this graph data as separate information (see \S \ref{section4:4.1}).

Before inputting into the model, the converted graph data are preprocessed by standardization and normalization. This step involves standardizing node positions $p_{n_{i}}$ and weather features $w$ within the batch, and normalizing the edge distance feature to facilitate the model's fitting during training and computation during prediction. Leveraging the Graph Attention Transformer (GAT), we integrate dropout and batch normalization to mitigate overfitting. The prediction confidence for each seed evolving into an error scenario is derived via a sigmoid layer.

The model is trained with historical data increments of $tr=1000$, splitting 80\% for training and 20\% for validation over 1000 epochs. During the prediction phase, the batch size fed into the model is equivalent to $N_m$, which represents the count of mutated seeds. We take into seeds with prediction confidence exceeding 0.5 (i.e., seeds anticipated to be potential error scenarios). These seeds are cherry-picked in a descending order based on confidence, with the highest $N_e$ seeds being selected for the ensuing execution.

\subsection{Execution and Result Monitoring}
\label{section4:6}

The chosen $N_e$ scenario seeds are used to set up the simulation environment and the system under test is integrated. The system's state data during the test within this scenario is also collected.

\textbf{Misbehavior detector.} This module is responsible for dynamically identifying improper behaviors exhibited by the ADS during testing. Error scenarios are defined as those in which the system under test exhibits one or more of the following erroneous behaviors:

\begin{itemize}
			[
		before=\vspace{-5pt}, 
		after=\vspace{-5pt} 
		]
    \item \textbf{Crash:} Collisions are detected when the vehicle's physical model contacts another entity. This process utilizes CARLA's collision detector APIs.
    \item \textbf{Red:} Running Red Lights is identified when the vehicle crosses an intersection against a red traffic signal, employing positional checks relative to traffic lights.
    \item \textbf{Speeding:} Speeding occurs when the vehicle's speed exceeds the CARLA-defined speed limit for a given road segment, with this limit serving as a fixed threshold.
	\item \textbf{Lane invasion:} Improper Lane Changes are detected through CARLA's lane invasion APIs, which trigger upon the vehicle crossing road lane markers.
    \item \textbf{Stuck:} Stagnation is defined when the vehicle remains stationary for a set duration, typically over five minutes, indicating a deadlock or system failure.

\end{itemize}

The misbehavior detector halts the simulation and the current seed fuzzing process immediately if any of these behaviors occur, classifying the currently executed scenario seed as an error scenario. Each error type's detection mechanism is parameterized to efficiently flag deviations from normal operational parameters and is designed to be indicative of potential ADS failures.

\textbf{Driving score calculator.} If ADS does not display improper behavior throughout the simulation, state information is collected from the back-end upon completion. This data, encompassing vehicle speed, braking signals, and acceleration, serves as the basis for calculating the ADS driving score, reflecting the vehicle's stability during operation, and indirectly indicating the extent to which the current scenario impacts this stability. We utilize the formula suggested by Drivefuzz for this calculation, with further details available in the supplementary material.

\vspace{-5pt}
\subsection{Seed Scheduling Strategy}
\label{section4:7}

After each fuzz generation round, we evaluate seeds by driving scores. If no error scenarios arise, the seed with the lowest score progresses to the next fuzzing round. Mutations then apply the random nearest neighbor strategy based on the seed's status to enhance error scenario discovery. Upon identifying an error scenario, the fuzz phase stops, and the original seed may re-enter the queue, subject to roulette-based frequency checks to prevent local optima and ensure diversity. If no errors occur, fuzzing proceeds with subsequent seeds until the queue depletes, then replenishes from the seed pool to maintain diversity.

This approach maintains a steady flow of new seeds for fuzzing, prevents stagnation at local optima, and ensures scenario variety.

\vspace{-5pt}
\subsection{Replay and Analysis of Test Results}
\label{section4:8}

In the error scenarios, we preserve the frame-by-frame driving trajectory coordinates of the ego car and surrounding objects throughout the testing phase, along with details such as weather parameters and object style types. Furthermore, we chronicle a gamut of data produced by the ADS in operation, including prediction and log records, ensuring the reproducibility of each scenario.

\begin{figure}[h!]
	\begin{center}
		\includegraphics[width=0.45\textwidth]{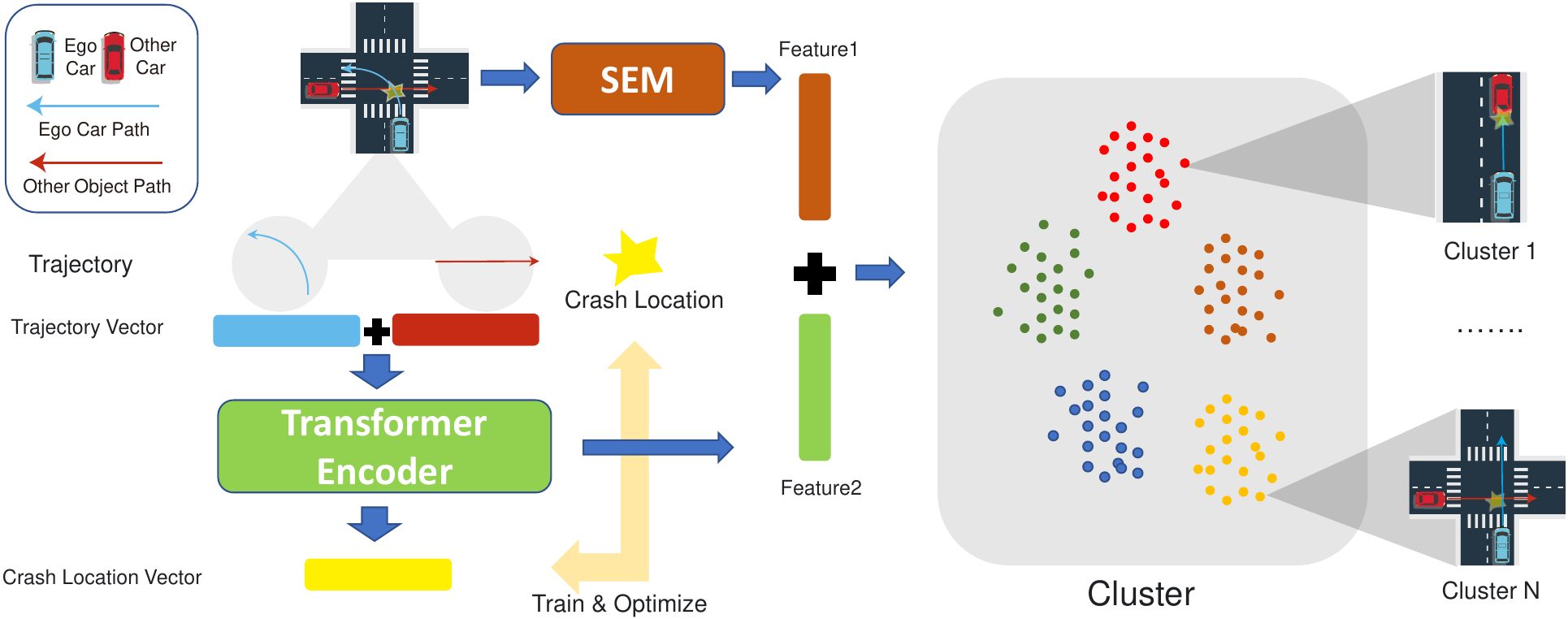}
		 \vspace{-10pt}
		\caption{Schematic of the vehicle condition clustering method based on self-supervised learning and SEM features.}
		\label{fig5}
		 \vspace{-15pt}
	\end{center}
\end{figure}

Considering the multitude of error scenarios unveiled, categorizing and organizing them is crucial. For collision incidents, we implement a feature extraction technique to reduce manual sorting efforts. This approach leverages features from the SEM and a self-supervised extractor based on the trajectory shared between the ego car and the object involved in the collision. These features are then amalgamated for clustering, aiming to distill characteristic vehicle interaction patterns, as illustrated in Figure \ref{fig5}. The manual review follows the clustering to confirm the uniqueness and significance of each pattern, especially since automated clusters can vary in utility and coherence. This step involves manually verifying and de-duplicating data, ensuring that the distilled patterns accurately reflect real-world interactions.

When dealing with incidents like speeding, traffic light violations, unlawful lane shifts, and stagnation, we examine the coordinates pinpointing these errors and classify them, extracting representative scenarios. Upon identifying these scenarios, we trace back system bugs by replaying the test sequence and meticulously analyzing the logs.

\section{Evaluation}
\label{section5}

To evaluate the performance of our framework, we conducted experiments to answer the following research questions:

\begin{itemize}
\item RQ1: How do components like mutation strategy and seed filtering impact the testing framework's performance?
\item RQ2: What is the accuracy and generalization capability of the SEM, and does it improve with more data?
\item RQ3: How does our method's capability in uncovering erroneous scenarios on a single seed compare to other methods?
\item RQ4: Does our method demonstrate improved efficiency in error scenario discovery and code coverage compared to Drivefuzz, and what role does the corpus play?
\item RQ5: Can the method detect bugs in ADS, and what are the common errors and collision-prone scenarios it reveals?
\end{itemize}

We address these questions through a series of detailed experiments, designed to explore the core aspects of our testing framework. The results and analyses of these experiments are presented in the following subsections, setting the stage for a comprehensive evaluation of our framework's performance and capabilities.

\subsection{Experimental Setup}
\label{section5:1}

\textbf{Experiment Configuration}: We configured the experiment parameters as follows: $Nc = 3$ for the maximum number of cycles, $N_m = 100$ as the mutation seed count (defaulting to $N_m = 3$ if the scenario evaluation model is unset due to omission of the filtering process), and $N_e = 3$ for the execution seed number. Details regarding the distribution of each type of road in the scenario seed corpus (from \S \ref{section4:2}) across the map can be found in the supplementary material. For our experimental evaluation, we randomly selected an initial seed from the scenario seed corpus spanning maps Town01 to Town05 within the CARLA. This selection strategy facilitates a comprehensive comparison across a wide range of scenarios, without the need for specific constraints on road types or traffic signs.

\textbf{ADS Types Studied}: Our study encompasses a range of ADS types: Autoware\footnote{\url{https://github.com/Autoware-AI},Version 1.14}\cite{autowarefoundation2023}, LAV\footnote{\url{https://github.com/dotchen/LAV}}\cite{chen2022lav} , Transfuser\footnote{\url{https://github.com/autonomousvision/transfuser}}\cite{Chitta2022PAMI}, and NEAT\footnote{\url{https://github.com/autonomousvision/neat}}\cite{Chitta2021ICCV} . Additionally, we evaluate a distinct ADS kind, termed Autopilots, which relies solely on planning and control modules, interfacing directly with the world state and map data in CARLA. While this type is rare in real-world applications, it offers valuable insights for validating our proposed testing methodologies. We include two logic-based Autopilot systems — Basic agent and Behavior agent, both formerly examined in the Drivefuzz project. Our study introduces three academic ADSs — LAV, Transfuser, and NEAT — all rigorously assessed via the CARLA Leaderboard.

\textbf{Baseline Comparisons}: Considering the interdependencies a\-mong CARLA, ADS, and the test method framework, Drivefuzz \cite{Kim22DriveFuzz} exclusively meets our needs for fuzzing from scratch and complete scenario construction where there are no predefined scenario sources. For approaches with predefined scenario sources, we have tailored these to our framework. We chose AV-Fuzzer \cite{9251068} and SAMOTA \cite{haq2022efficient} for demonstrating our performance comparisons on a single seed across three road types of scenarios at the same geographical locations.

\subsection{Results}
\label{section5:2}

\subsubsection{Component Impact on Performance}
\label{section5:2:1}

To compare the roles of the two-stage mutation strategy and SEM in the entire framework, we conducted relevant ablation experiments. These are divided into: 1) RMS: using random mutation; 2) 2SMS: using a two-stage mutation strategy; 3) RMS+SEM: using random mutation followed by SEM filtering; 4) 2SMS+SEM: using a two-stage mutation strategy followed by SEM filtering, which is the complete method. We tested the Autoware for 6 hours using these various combinations. As Autoware is a mature ADS already applied in the industry, the testing efficiency performance evaluated is relatively objective.

\begin{figure}[h!]
	\centering
	\subfloat[Components]{\includegraphics[width=0.23\textwidth]{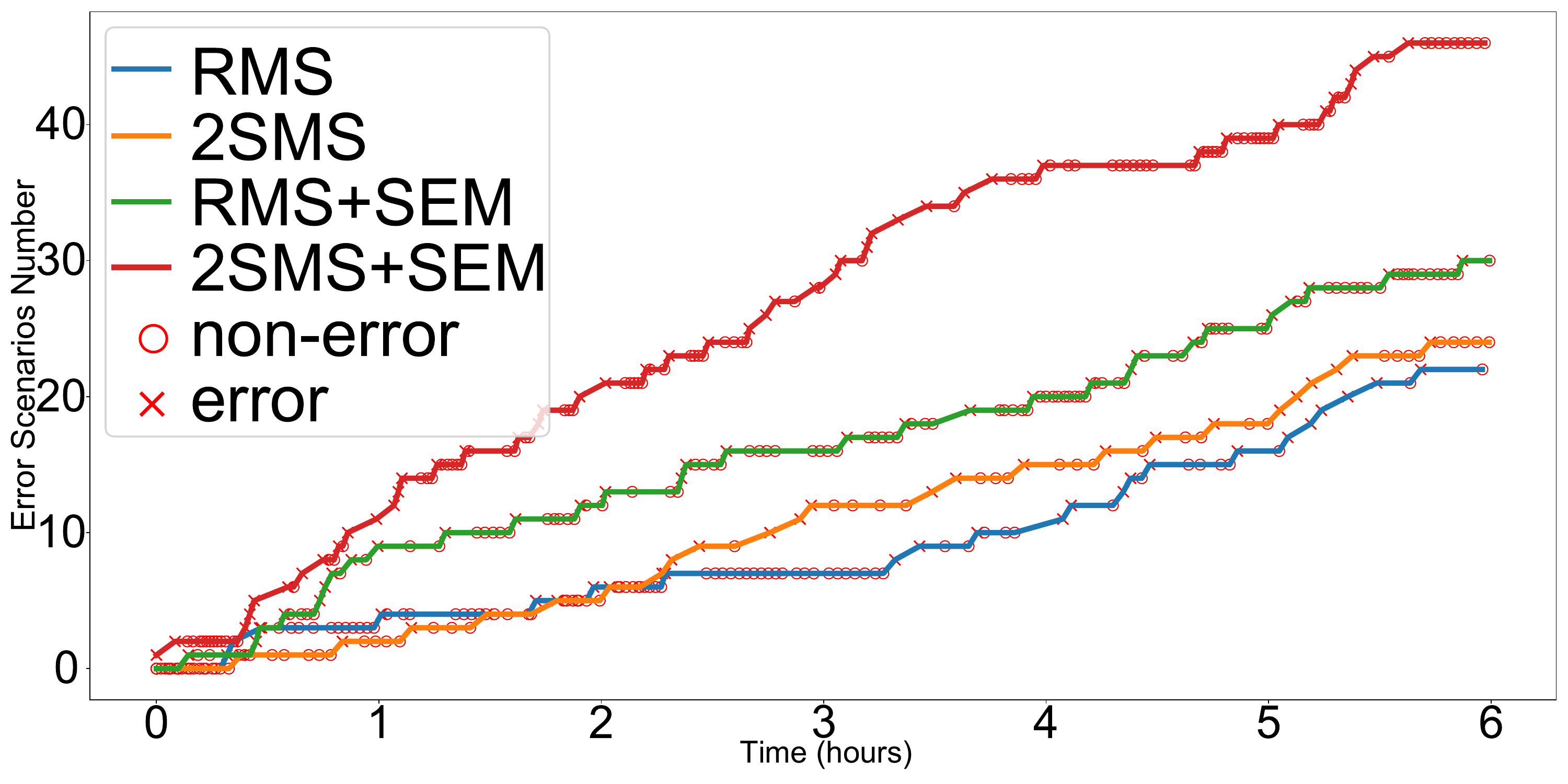}\label{fig6-1}}
	\subfloat[SEM under different data sizes]{\includegraphics[width=0.23\textwidth]{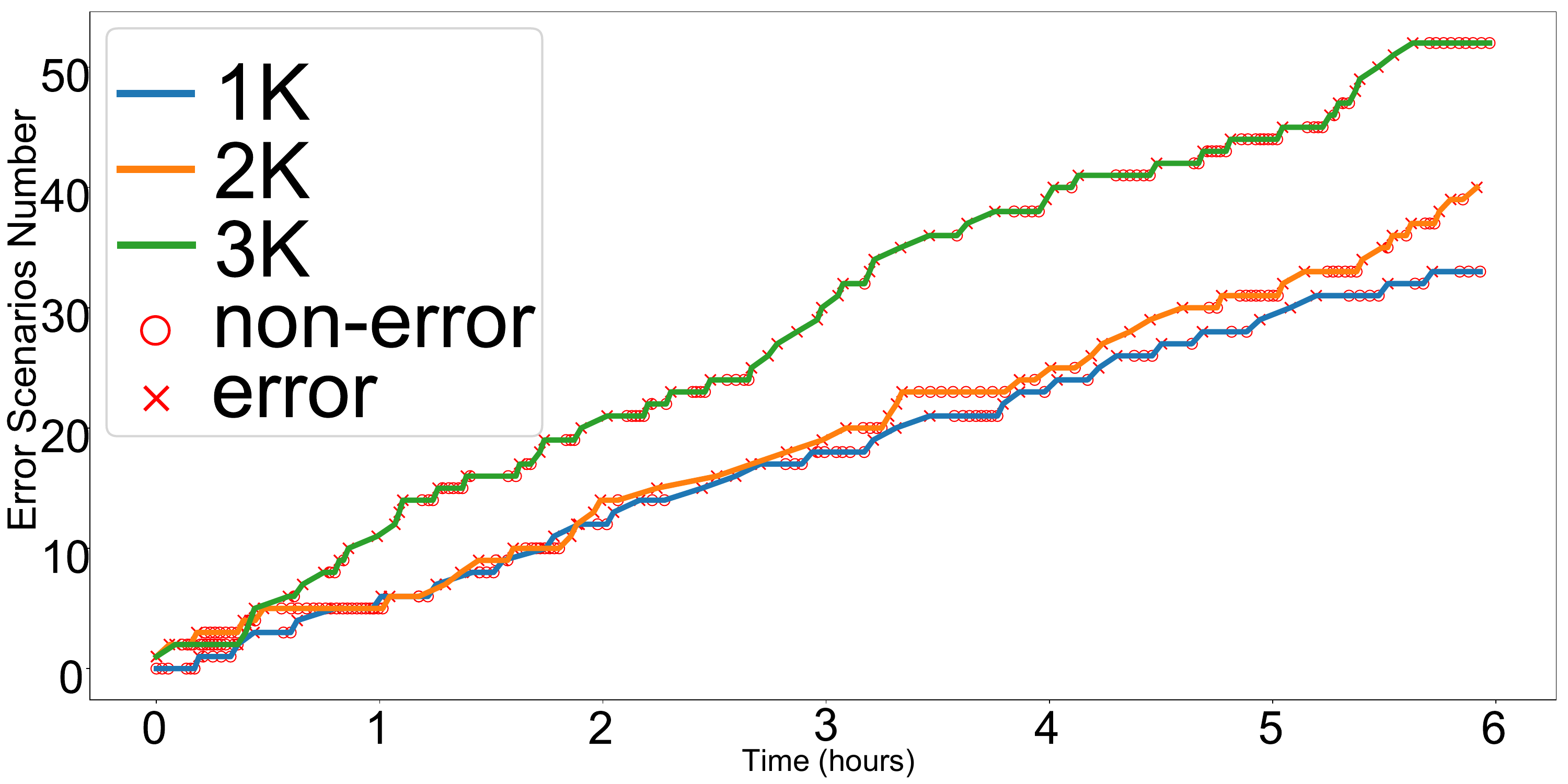}\label{fig6-2}}
	\vspace{-10pt}
	\caption{Efficiency comparison of error scenario generation under different configurations.}
	\label{fig6}
	\vspace{-15pt}
\end{figure}

From Figure \ref{fig6-1}, it can be observed that as the components are progressively refined, the efficiency of the framework in uncovering error scenarios also gradually improves. In the comparison of whether to adopt a two-stage mutation strategy, it is evident that the random mutation strategy tends to stagnate in terms of the number of scenarios uncovered during the exploration process (e.g., at the 3-hour mark). With the guidance of the driving score, the two-stage mutation strategy can alleviate such issues, with this adaptive strategy better delving into and further exploring the vicinity of error scenarios. In the comparison of whether to use SEM, it's clear that the efficiency is higher with SEM. SEM filters mutated seeds based on historical data, effectively selecting high-quality seeds for execution, thereby enhancing efficiency.

\vspace{-10pt}

\subsubsection{SEM's Accuracy and Growth}
\label{section5:2:2}

In the previous section, we discussed the impact of having or not having SEM on the testing framework. Next, we need to evaluate the growth performance of SEM under different amounts of historical data, as well as the accuracy and generalization of seed filtering capabilities across different testing systems. As described in \S \ref{section4:5}, we train SEM whenever the historical test data reaches a certain quantity and retain 20\% of the data for model evaluation. In our experiments, we updated the model three times when the historical data quantities reached 1k, 2k, and 3k, respectively.

We integrated the SEM from these three different stages into the testing framework and conducted a 6-hour test on Autoware. The efficiency of error scenario discovery is shown in Figure \ref{fig6-2}. As the volume of training historical data increased, SEM's filtering capability improved, driving an enhancement in discovery efficiency. The difference in efficiency between 1k and 2k data volumes was not particularly significant, but when the data volume reached 3k, there was a substantial increase in efficiency. This indicates that SEM has significant growth potential and can effectively improve the framework's discovery efficiency driven by historical data.

\begin{figure}[h!]
	\centering
	\subfloat[Accuracy/Loss]{\includegraphics[width=0.2\textwidth]{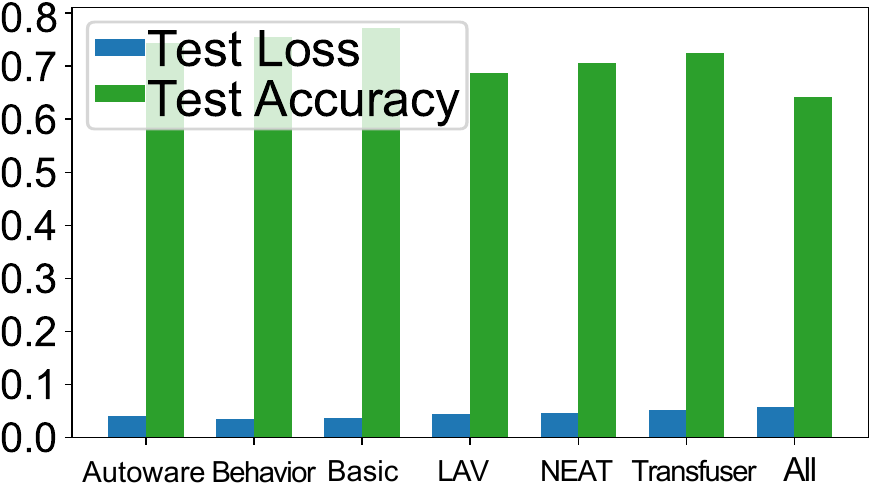}\label{fig7-1}}
	\subfloat[Precision/Recall]{\includegraphics[width=0.2\textwidth]{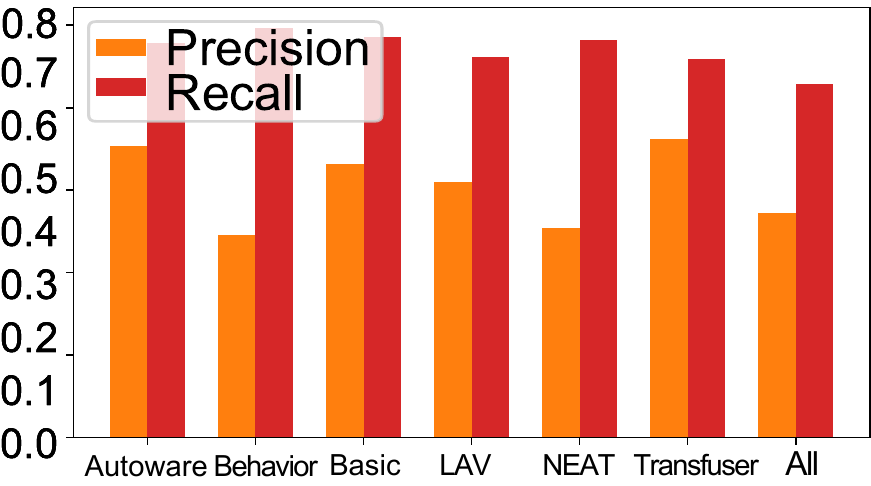}\label{fig7-2}}
	\vspace{-10pt}
	\caption{Performance of the SEM under different systems.}
	\label{fig7}
	\vspace{-15pt}
\end{figure}

Furthermore, we also had SEM evaluate the 20\% data during each of the three training updates, taking the average accuracy, loss, precision, and recall from the three tests. As shown in Figure \ref{fig7}, in terms of accuracy and loss, SEM effectively ensures good generalization across data generated from different systems and performs well in overall data. In terms of precision and recall, the model has a strong ability to capture error scenarios, striving not to miss any. 

While the model may yield false positives within normal scenarios, prioritizing the detection of even a minimal number of error scenarios is crucial within our work's context.

Detecting false negatives presents challenges due to the model's limitations and the intricate nature of error patterns in ADS, leading to some error scenarios being missed. Among these are: 1) errors occurring outside the expected scenario boundaries; 2) unique error types, such as stagnation, speeding, or failing to reach the intended destination accurately; and 3) the ego car not following the task route due to planning module issues, resulting in unexpected errors. These complex error patterns are difficult for the model to accurately represent and predict, requiring further advancements. Future research will focus on enhancing the model's structure for improved handling of complex conditions; enriching the input graph data with additional nodes and features for the precise depiction of complex trajectories and scenarios; and enlarging the dataset to let SEM better capture and predict a variety of complex error behaviors.

\vspace{-13pt}

\subsubsection{Single-Seed Discovery Performance}
\label{section5:2:3}

We conducted a six-hour Autoware system test across three types of road scenarios at the same geographical location, assessing each method's ability to generate erroneous scenarios.

\vspace{-13pt}
\begin{figure}[h!]
	\centering
	\subfloat[AV-Fuzzer]{\includegraphics[width=0.1\textwidth]{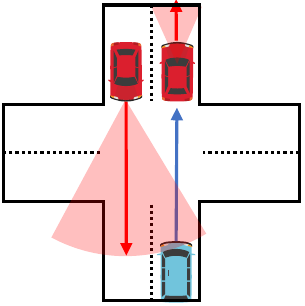}\label{seed-1}} 
	\subfloat[SAMOTA]{\includegraphics[width=0.1\textwidth]{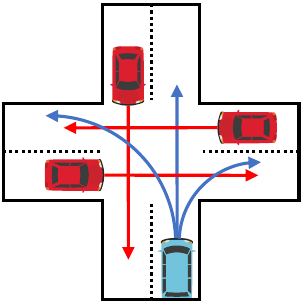}\label{seed-2}}
	\subfloat[Ours]{\includegraphics[width=0.1\textwidth]{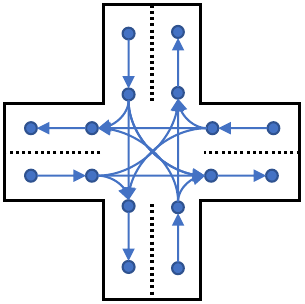}\label{seed-3}}
	\vspace{-10pt}
	\caption{Initial seed information for three methods at a crossroad.}
	\vspace{-10pt}
	\label{initial-seed}
\end{figure}

In which, AV-Fuzzer determines the initial positions of all vehicles in a given scenario and mutates the maneuvers of other vehicles. SAMOTA optimizes scenario parameter vectors, which are used to select from three predefined ego car tasks and determine whether to position other vehicles in one of three types of locations. Our method, leveraging automatically generated corpus information, allows the mutator to select appropriate positions and paths for entities, creating scenarios with various interactions, as shown in Figure \ref{initial-seed}. Furthermore, compared to the two methods, our approach introduces additional variation techniques not present in AV-Fuzzer and SAMOTA. These include placing pedestrians within scenarios, altering vehicle appearances, and adding environmental challenges such as water puddles. In terms of weather conditions, while SAMOTA limits choices to eight predefined options, our method allows for a broader range of weather variability.

\textbf{Total Error Scenarios Results}: As shown in Table \ref{tab0}, on crossroad and T-intersection, our method was able to identify a higher number of erroneous scenarios, the detailed reasons for which will be discussed subsequently. On the straight road, however, the number of erroneous scenarios uncovered by our method was lower than that of AV-Fuzzer. This can be primarily attributed to the fact that the Autoware system, on the CARLA platform, does not recognize any traffic lights (as noted in Autoware Bug9 in the supplementary material Table 2), leading to frequent red light violations associated with the predefined mission paths. Secondly, AV-Fuzzer is specifically designed for generating malicious merging trajectories of other vehicles on straight roads, resulting in numerous repetitive scenarios like those depicted in Figure \ref{fig13}, S6 and S7. Although our method generated fewer scenarios than AV-Fuzzer on straight roads, the variety of scenarios was more diverse.

\vspace{-10pt}
\begin{table}[h!]
	\caption{Comparison of error scenarios for three methods with different road types}
	\vspace{-10pt}
	\label{tab0}
	\centering
	\scalebox{0.65}{
		\begin{tabular}{ccccccp{4cm}}
			\toprule
			\textbf{Type} &                      
			\textbf{Method}    & 
			\textbf{Count} & 
			\textbf{Crash}              & 
			\textbf{Stuck}              & 
			\textbf{Red}                & 
			\textbf{Details} \\
			\midrule
			\multirow{3}{*}{StraightRoad}  & AV-Fuzzer    & \textbf{60}           & 32    & 0     & 28  & S6, S7                                           \\
			& SAMOTA       & 9                     & 3     & 3     & 3   & S13                                               \\
			& ScenarioFuzz & 33                    & 12    & 2     & 19  & S3, S8, S9, S11, S13                                          \\
			\midrule
			\multirow{3}{*}{CrossRoad}     & AV-Fuzzer    & 0                     & 0     & 0     & 0   & -                                                         \\
			& SAMOTA       & 12                    & 1     & 0     & 11  & S16                                                       \\
			& ScenarioFuzz & \textbf{25}           & 8     & 14    & 3   & \parbox{4cm}{S7, S18, S25, S29(child, lying)}     \\
			\midrule
			\multirow{3}{*}{T-intersection} & AV-Fuzzer    & 4                     & 4     & 0     & 0   & S54                                                       \\
			& SAMOTA       & 5                     & 4     & 1     & 0   & S32                                                       \\
			& ScenarioFuzz & \textbf{40}           & 22    & 18    & 0   & \parbox{4cm}{S34, S35(adult, child), S36(adult, child), S37, S38, S39, S40, S49} \\
			\bottomrule                  
		\end{tabular}
	}
	\vspace{-10pt}
\end{table}

\begin{figure}[h!]
	\centering
	\subfloat[Straight Road]{\includegraphics[width=0.39\textwidth]{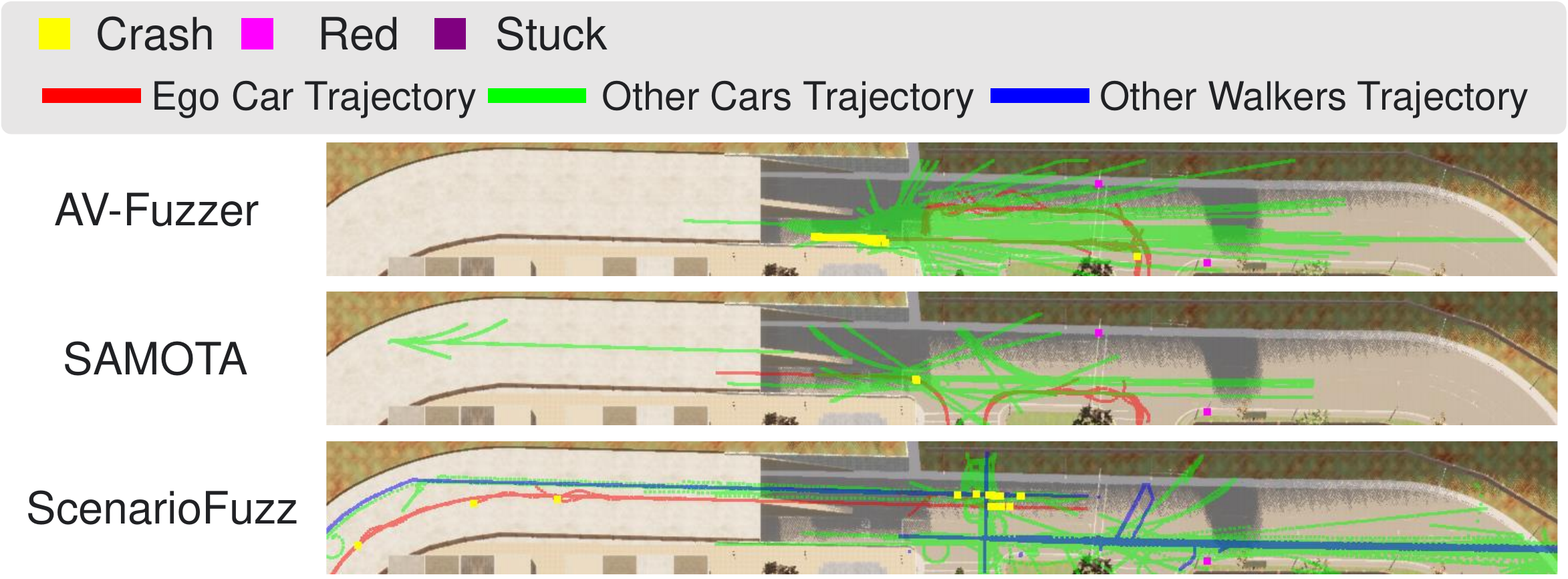}\label{traj-1}} \\
	\vspace{-10pt}
	\subfloat[Cross Road]{\includegraphics[width=0.39\textwidth]{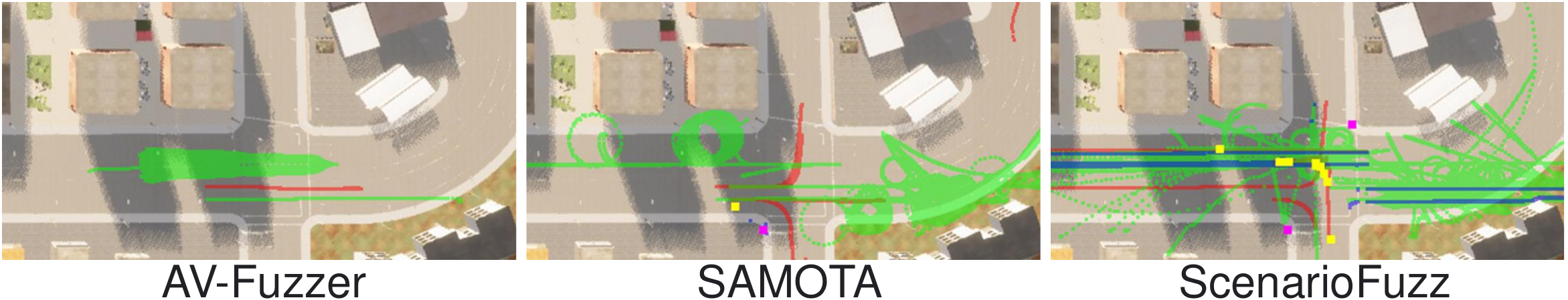}\label{traj-2}} \\
	\vspace{-10pt}
	\subfloat[T-intersection]{\includegraphics[width=0.39\textwidth]{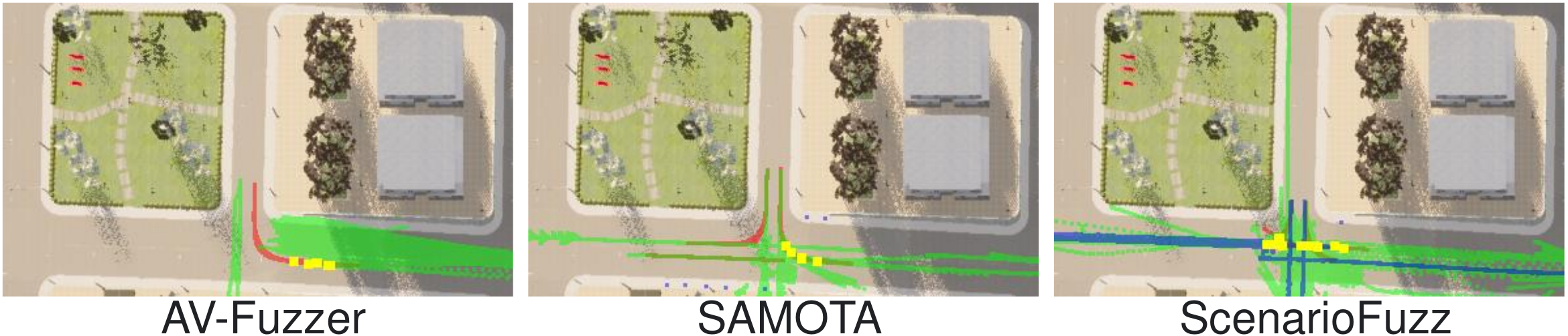}\label{traj-3}}
	\vspace{-10pt}
	\caption{Comparative Analysis of Scenario Trajectories.}
	\label{traj-single-seed}
\vspace{-10pt}
\end{figure}

\textbf{Scenario Trajectory Distribution and Method Analysis}: To assess the influence of different scenario generation processes, we visualized the trajectories of both the ego car and other objects for all three methods, pinpointing collision sites and other errors as demonstrated in Figure \ref{traj-single-seed}.

AV-Fuzzer and SAMOTA similar to many existing methods, predefine the positions and paths for the ego car and other objects, our approach diverges significantly. Specifically, AV-Fuzzer focuses on controlling the maneuver of other objects, while SAMOTA emphasizes combinations of ego car tasks with other vehicles' presences within the same scenario. In contrast, our method employs an automated map crawling technique to extract all potential placement points and paths from the road network, thereby furnishing both the ego car and other objects with comprehensive path and placement information from our initial seed, as depicted in Figure \ref{initial-seed}.

The initial seed setup of our approach addresses the limitations associated with manual predefined settings. Whereas AV-Fuzzer and SAMOTA require manually setting within the map, getting the limited quantity of initial seeds, our method automates the extraction of segment information across different geographical locations and road types, efficiently using the resources of the map. This automation expands the testing breadth and allows for a richer set of scenario settings, enabling thorough exploration of unforeseen ADS errors by generating a wider range of scenarios and interactions, ensuring a more extensive validation process. 

The completeness of the initial seed significantly influences the diversity of the resultant scenarios. The details in Table \ref{tab0} illustrate the variety of collision scenarios generated by each method across different road types (with scenario IDs corresponding to those in Figure \ref{fig13}). Our approach yields a more diverse array of outcomes compared to the others. For instance, on straight roads, AV-Fuzzer manipulates other vehicles' maneuvers, leading predominantly to collisions caused by malicious merging due to fixed initial positioning. This limitation hinders its performance in more complex scenarios such as crossroads and T-intersections, where the space for maneuver generation does not intersect with the ego car's task trajectory. Conversely, while SAMOTA sets more route options for different road types, its reliance on CARLA's built-in Autopilot for other vehicles results in unrealistic behaviors, such as driving in circles, due to inherent planning issues in Autopilot, thus generating many insignificant scenarios. Our method's comprehensive initial seed expands the variety of generated scenarios, beyond the constraints of predefined behavior sets, enhancing the coverage of trajectory distributions and outperforming the previous methods, especially as shown in Figure \ref{traj-single-seed}.

\vspace{-15pt}

\vspace{-8pt}
\begin{table}[ht]
	\centering
	\caption{Model Performance Comparison}
	\vspace{-12pt}
	\scalebox{0.75}{
		\begin{tabular}{lcc}
			\hline
			\textbf{Model} & \textbf{Pearson Correlation Coefficient $\uparrow$} & \textbf{Brier Score $\downarrow$} \\
			\hline
			RBFN            & 0.0578             & 0.8142          \\
			SEM             & \textbf{0.2965}    & \textbf{0.2748}          \\
			\hline
	\end{tabular}}
	\label{tab:model_performance}
\end{table}

\textbf{Filter Model Performance}: The SAMOTA method employs a combination of global and local search techniques, incorporating a Radial Basis Function Network (RBFN) for population fitness prediction and filtering. However, given that RBFN is a regression model rather than a classification model, we opted to forgo accuracy as a performance metric. Instead, we assessed the performance difference between RBFN and our SEM by calculating the correlation between the model's output and actual test results, like the Pearson correlation coefficient \cite{turney-2023} and Brier score \cite{brier1950verification}. Our SEM outperformed the RBFN, which is merely a 3-layer fully connected network responsible for predicting based on abstracted scenario parameter vectors. This simple representation struggles to encapsulate the complexities of dynamic environments and the uncertainties inherent in testing processes, indicating that our SEM is an effective filtering mechanism. As shown in Table \ref{tab:model_performance}.

\subsubsection{Drivefuzz Comparison and Corpus Role}
\label{section5:2:4}

In assessing generation efficiency, we conducted tests on six ADS systems using three different approaches: Drivefuzz, Drivefuzz complemented with our corpus, and our proprietary method, ScenarioFuzz. Each system underwent five testing iterations. We then computed the average quantity of erroneous scenarios generated (depicted as lines) and their standard deviation (illustrated as areas) during these tests\footnote{With the exception of NEAT, which was tested for six hours, all systems underwent the same duration of testing. NEAT required extended execution times due to computational demands, resulting in a prolonged testing period of 12 hours to guarantee an adequate number of executions.}.

\begin{figure}[htbp]
	\centering
	\subfloat[Autoware]{\includegraphics[width=0.15\textwidth]{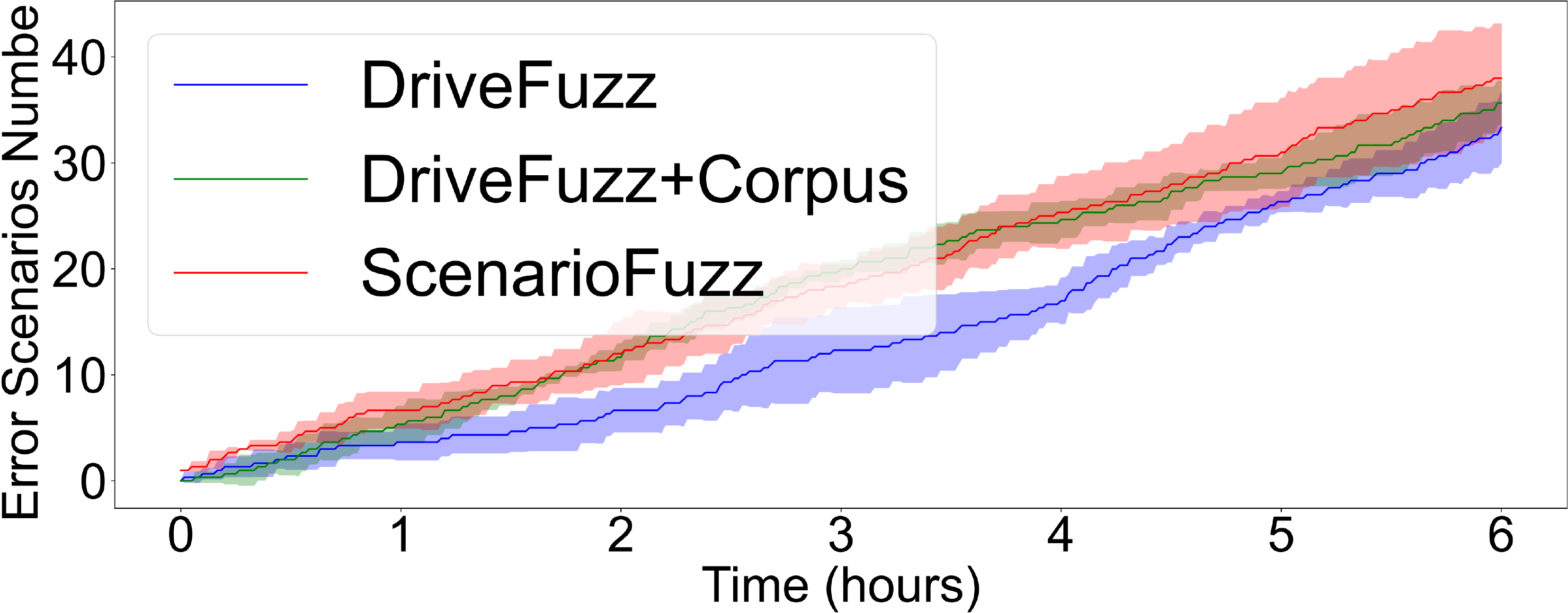}\label{fig8-1}}
	\hspace{0.08cm}  
	\subfloat[Basic]{\includegraphics[width=0.15\textwidth]{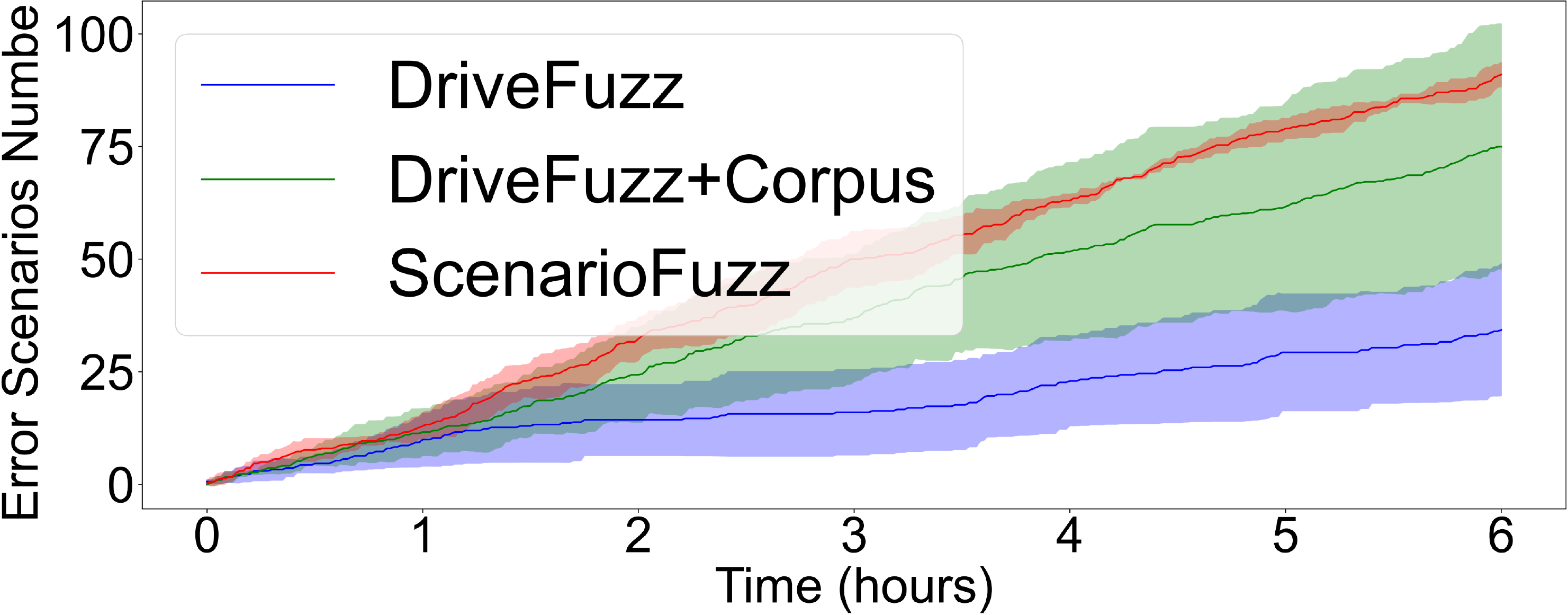}\label{fig8-2}}
	\hspace{0.08cm}  
	\subfloat[Behavior]{\includegraphics[width=0.15\textwidth]{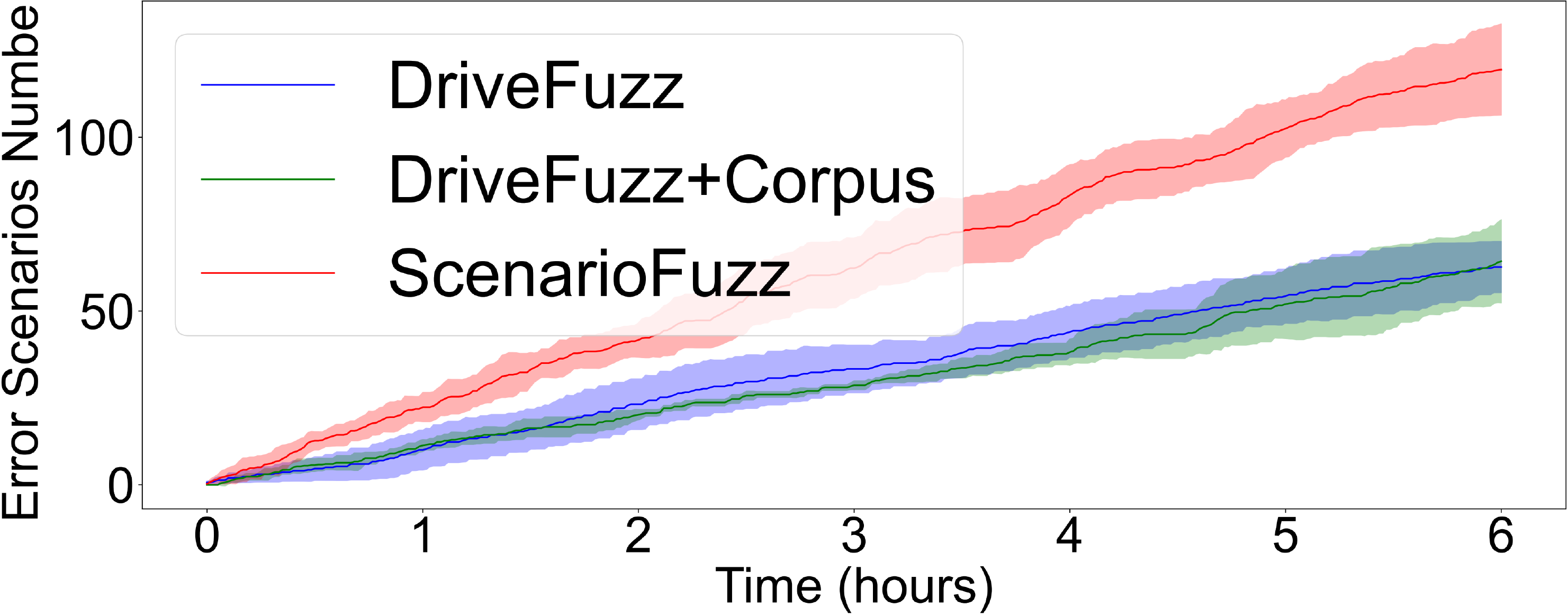}\label{fig8-3}} \vspace{-10pt}\\	
	
	\subfloat[LAV]{\includegraphics[width=0.15\textwidth]{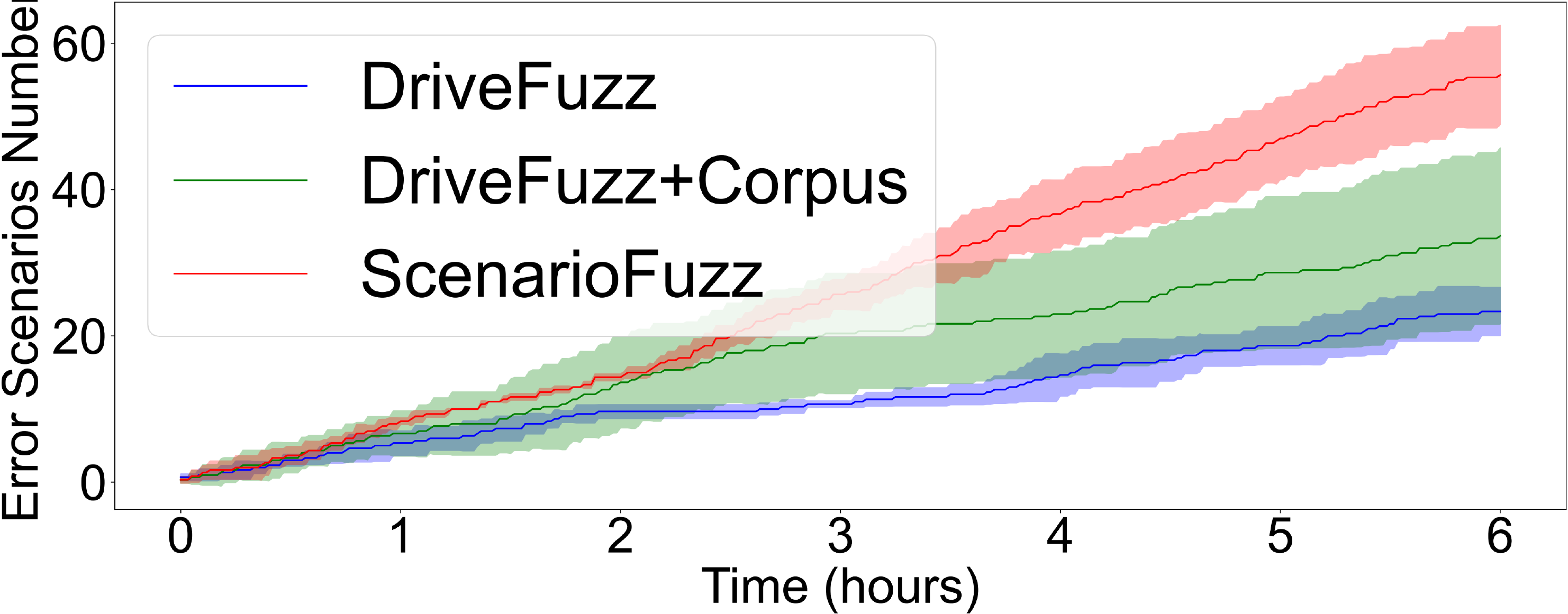}\label{fig8-4}}
	\hspace{0.08cm}  
	\subfloat[NEAT]{\includegraphics[width=0.15\textwidth]{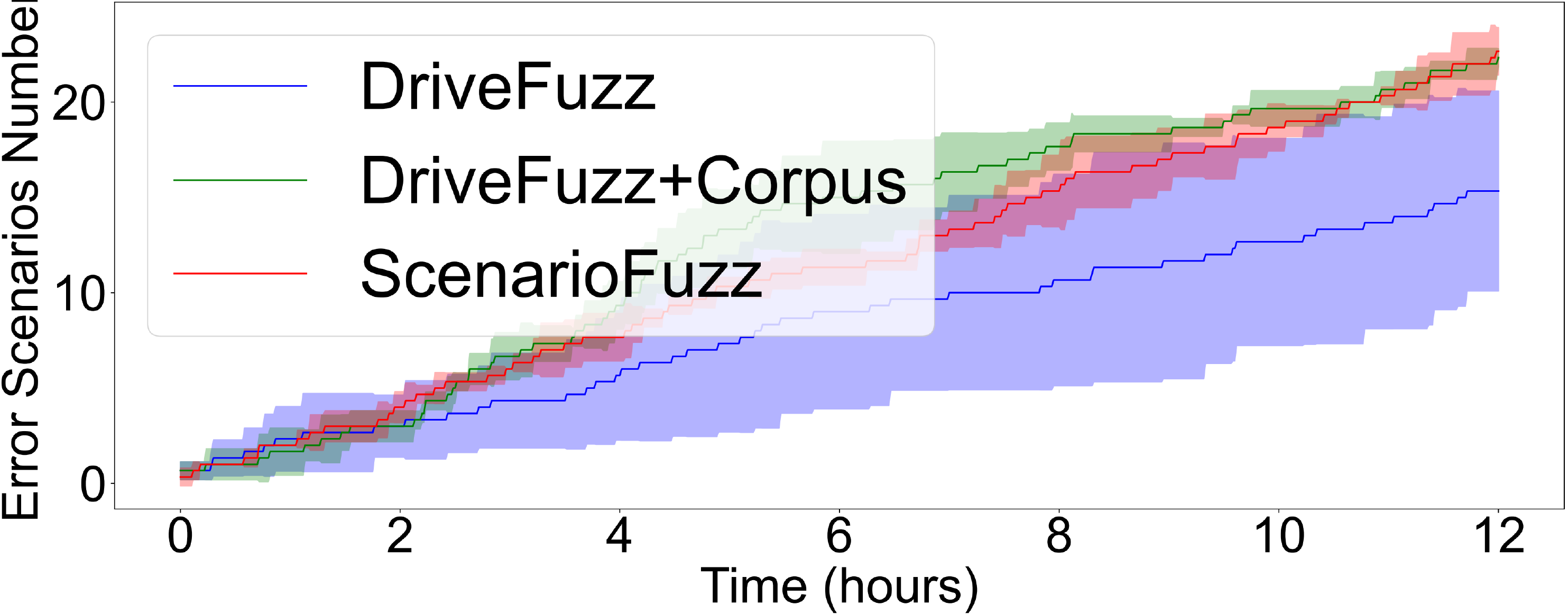}\label{fig8-5}}
	\hspace{0.08cm}  
	\subfloat[Transfuser]{\includegraphics[width=0.15\textwidth]{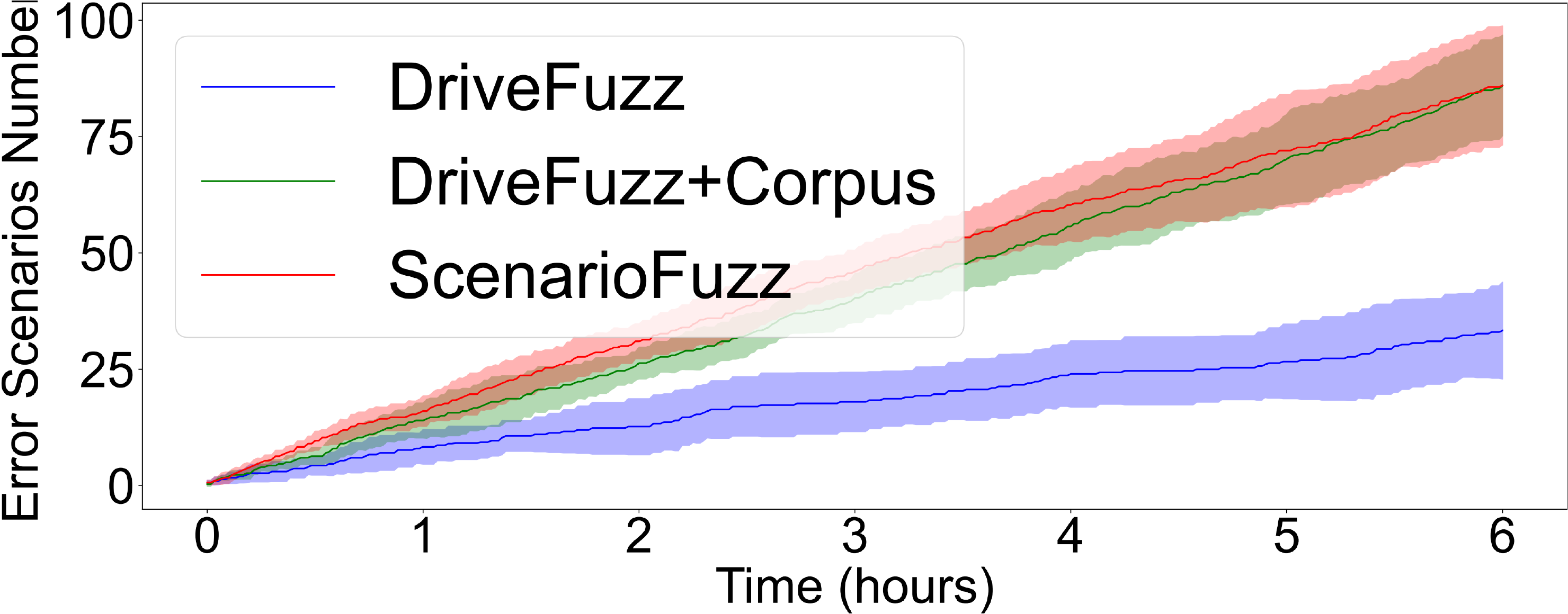}\label{fig8-6}}  \vspace{-10pt}
	
	\caption{Comparison of efficiency in generating error scenarios across various systems using the two methods.}
	\label{fig8}
	\vspace{-10pt}
\end{figure}

\textbf{Efficiency}: As can be seen from Figure \ref{fig8}, our method consistently outperforms Drivefuzz in efficiency across all systems, and the stability across the five tests is also notably superior. The magnitude of difference between the two methods on different systems is also related to the error patterns inherent in the systems. On Autoware, although the overall number advantage isn't pronounced, it's mainly because the Autoware system struggles to effectively recognize traffic lights in CARLA. Drivefuzz, which randomly selects start and end points on the map, predominantly captures red-light running scenarios. NEAT, due to performance issues, frequently stagnates during testing, which severely hampers the potential to unearth other types of errors.

\vspace{-15pt}
\vspace{-5pt}

\textbf{Execution Time and Number of Error Scenarios}: In \S \ref{section1}, we emphasized that an overly extensive 'stage scope' can significantly increase the execution time for each seed. This, within a limited testing duration, reduces the number of seed mutations and executions, hindering a more in-depth testing and error scenario discovery. We calculated the average seed execution duration\footnote{excluding scene construction time, which for both methods typically ranged between 5-25 seconds} and the average number of error scenarios discovered at the end of the five tests, as shown in Table \ref{tab1}. Compared to Drivefuzz, our method reduced the execution time by an average of 60.3\% and discovered an average of 103\% more error scenarios.

\begin{table}[h!]
	
	\caption{Comparison of execution costs and error scenarios.}
	\vspace{-10pt}
	\label{tab1}
	\centering
	\scalebox{0.65}{
		\begin{tabular}{ccccccc}
			\toprule
			\textbf{ADS} &
			\textbf{Method} &
			\textbf{\footnotesize\begin{tabular}[c]{@{}c@{}}Execution\\time(s)\end{tabular}} &
			\textbf{\footnotesize\begin{tabular}[c]{@{}c@{}}Increase\\rate(\%)\end{tabular}} &
			\textbf{\footnotesize\begin{tabular}[c]{@{}c@{}}Error\\count\end{tabular}} &
			\textbf{\footnotesize\begin{tabular}[c]{@{}c@{}}Increase\\rate(\%)\end{tabular}} &
			\textbf{\footnotesize\begin{tabular}[c]{@{}c@{}}Execution\\count\end{tabular}} \\
			\midrule
			\multirow{3}{*}{Autoware}   & DriveFuzz           & 393.17  &       & 32           &        & 61  \\
			                            & DriveFuzz+Corpus    & 279.7	& 28.86	& 35	       & 9.38	& 78  \\
										& ScenarioFuzz        & 277.98  & 29.30 & \textbf{38}  & 18.75  & 72  \\
			\midrule
			\multirow{3}{*}{Basic}    & DriveFuzz           & 232.85  &       & 34           &        & 95  \\
			                          & DriveFuzz+Corpus    & 63.3	  & 72.82 & 75	         & 120.59 & 344 \\
			                          & ScenarioFuzz        & 64.23   & 72.42 & \textbf{90}  & 164.71 & 298 \\
			\midrule
			\multirow{3}{*}{Behavior} & DriveFuzz           & 247.17  &       & 62           &        & 100 \\
			                          & DriveFuzz+Corpus    & 72.8    & 70.55 & 65           & 4.84   & 303  \\
			                          & ScenarioFuzz        & 43.66   & 82.34 & \textbf{120} & 93.55  & 346 \\
			\midrule
			\multirow{3}{*}{LAV}      & DriveFuzz           & 625.07  &       & 23           &        & 40  \\
				                      & DriveFuzz+Corpus    & 296.51  & 52.56 & 33           & 43.48  & 75  \\
			                          & ScenarioFuzz        & 231.33  & 62.99 & \textbf{55}  & 139.13 & 88  \\
			\midrule
			\multirow{3}{*}{NEAT}     & DriveFuzz           & 1358.94 &       & 15           &        & 35  \\
			                          & DriveFuzz+Corpus    & 731.07  & 46.20 & 22           & 46.67  & 66  \\
			                          & ScenarioFuzz        & 558.64  & 58.89 & \textbf{22}  & 46.67  & 76  \\
			\midrule
			\multirow{3}{*}{Transfuser} & DriveFuzz          & 384.13  &       & 33           &        & 58  \\
			                            & DriveFuzz+Corpus       & 169.46  & 55.88 & 84           & 154.55  & 132  \\
			                            & ScenarioFuzz       & 170.31  & 55.66 &  \textbf{85}  & 157.58 & 113  \\
			\bottomrule
		\end{tabular}
		
	}
	\vspace{-10pt}
\end{table}

\begin{figure}[h!]
	\centering
	\includegraphics[width=0.45\textwidth]{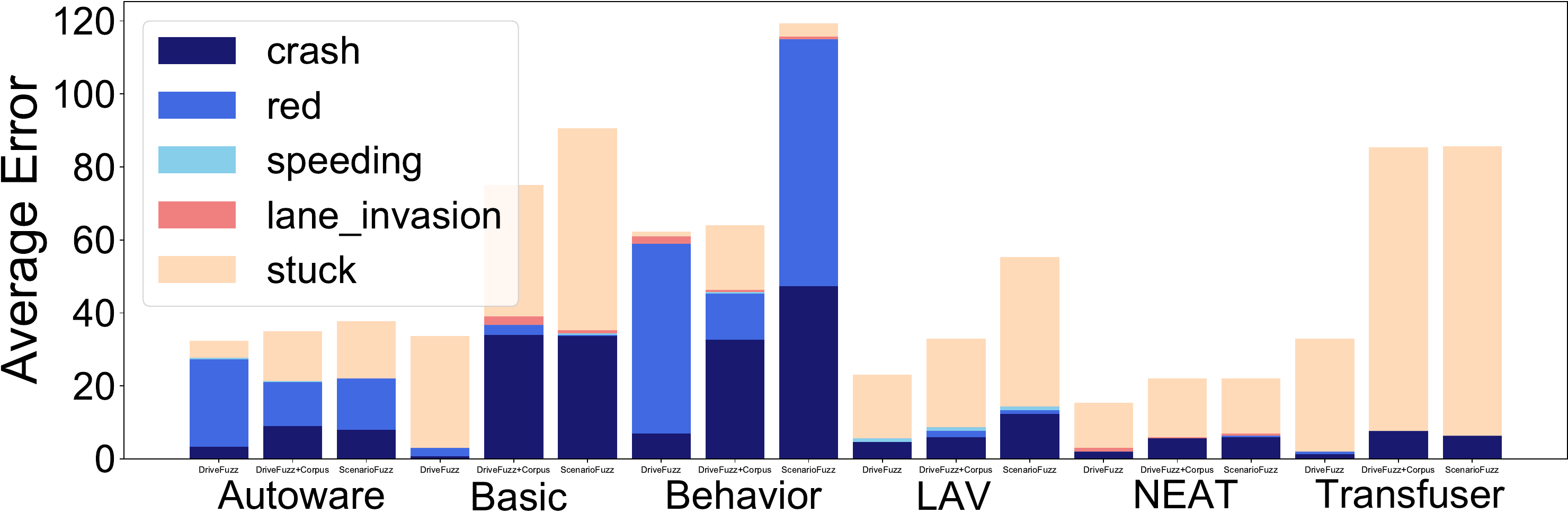}
	\vspace{-10pt}
	\caption{Count and types of error scenarios for two methods across various systems.}
	\label{fig9}
	\vspace{-10pt}
\end{figure}

\textbf{Error Scenario Types}: For the number of error types within the generated error scenarios, refer to Figure \ref{fig9}. From left to right, the order of the systems is consistent with the previous figure, with DriveFuzz on the left and ScenarioFuzz on the right for each system. The height of the bar chart indicates the average number of error scenarios discovered by the respective method for the given system, with different colors reflecting the proportion of each error type. The intensity of the color also indicates the severity of the error (refer to the order in related works \cite{abellan2013analysis,gong2021classification,SusanV}). In terms of error types, our method has a distinct advantage in discovering collision scenarios, thanks to the setup of the corpus (which will be detailed later) and the filtering capability of SEM. By transforming scenario data into graph-type data, which combines both static and dynamic information of the scenario, SEM effectively learns the respective representations from historical data, thereby filtering out scenario seeds with a higher potential for collisions for execution.

	\vspace{-10pt}

\textbf{Code Coverage}: In our testing process, we specifically measured the code line coverage for the perception, planning, and control modules of the Autoware system, and for the planning and control modules of the Basic and Behavior systems, utilizing lcov and the Python library 'coverage', respectively. By treating each generated scenario as a unique test case, we aimed to trigger diverse operational conditions within these modules. This approach allows us to assess how thoroughly different scenarios exercise the ADS's code, highlighting areas that remain untested. High coverage indicates that the system has been activated and executed across a diverse range of environments.

\begin{figure}[h!]
	\centering
	\subfloat[Autoware]{\includegraphics[width=0.15\textwidth]{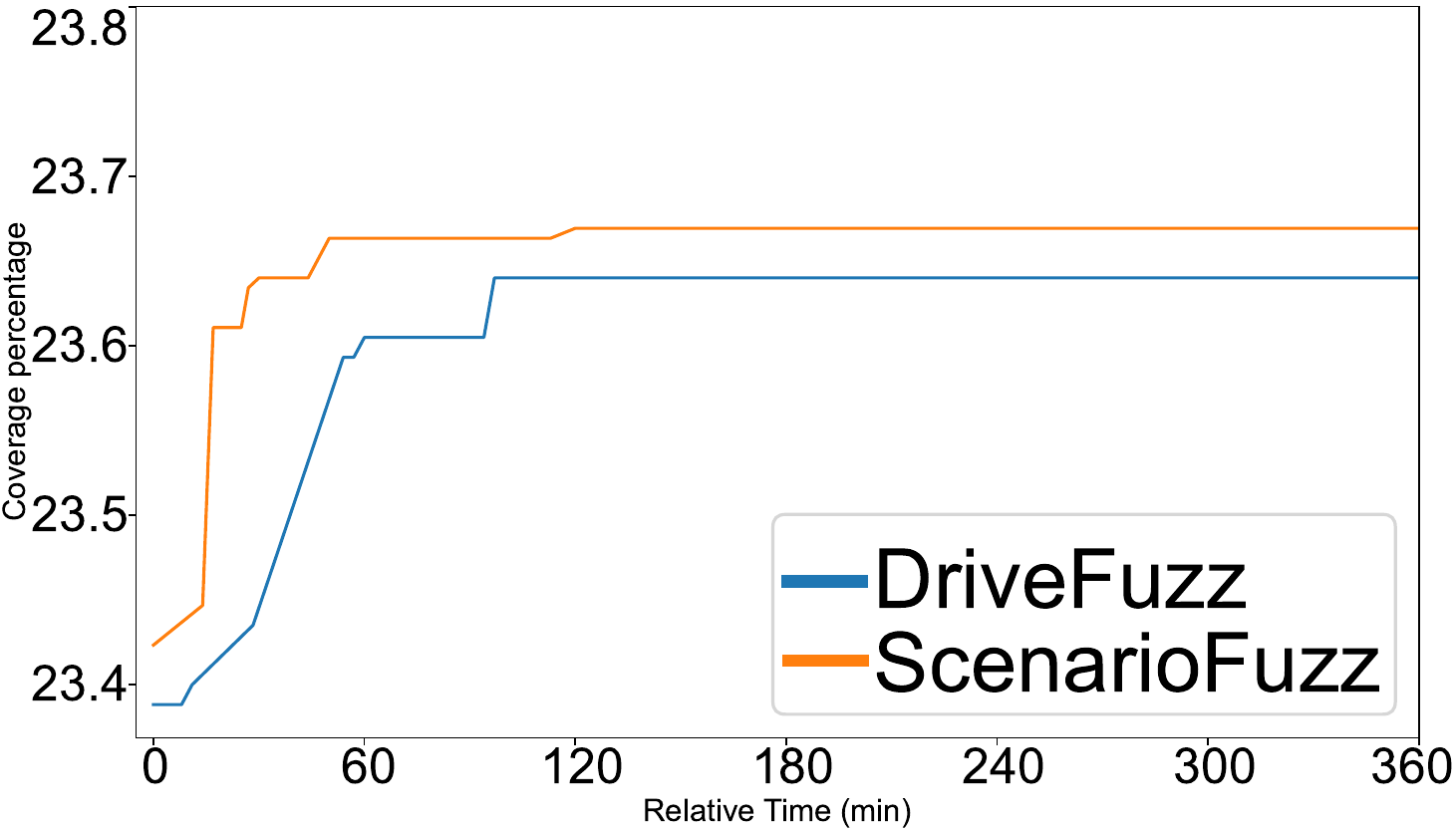}\label{fig10-1}}
	\hspace{0.08cm}  
	\subfloat[Basic]{\includegraphics[width=0.15\textwidth]{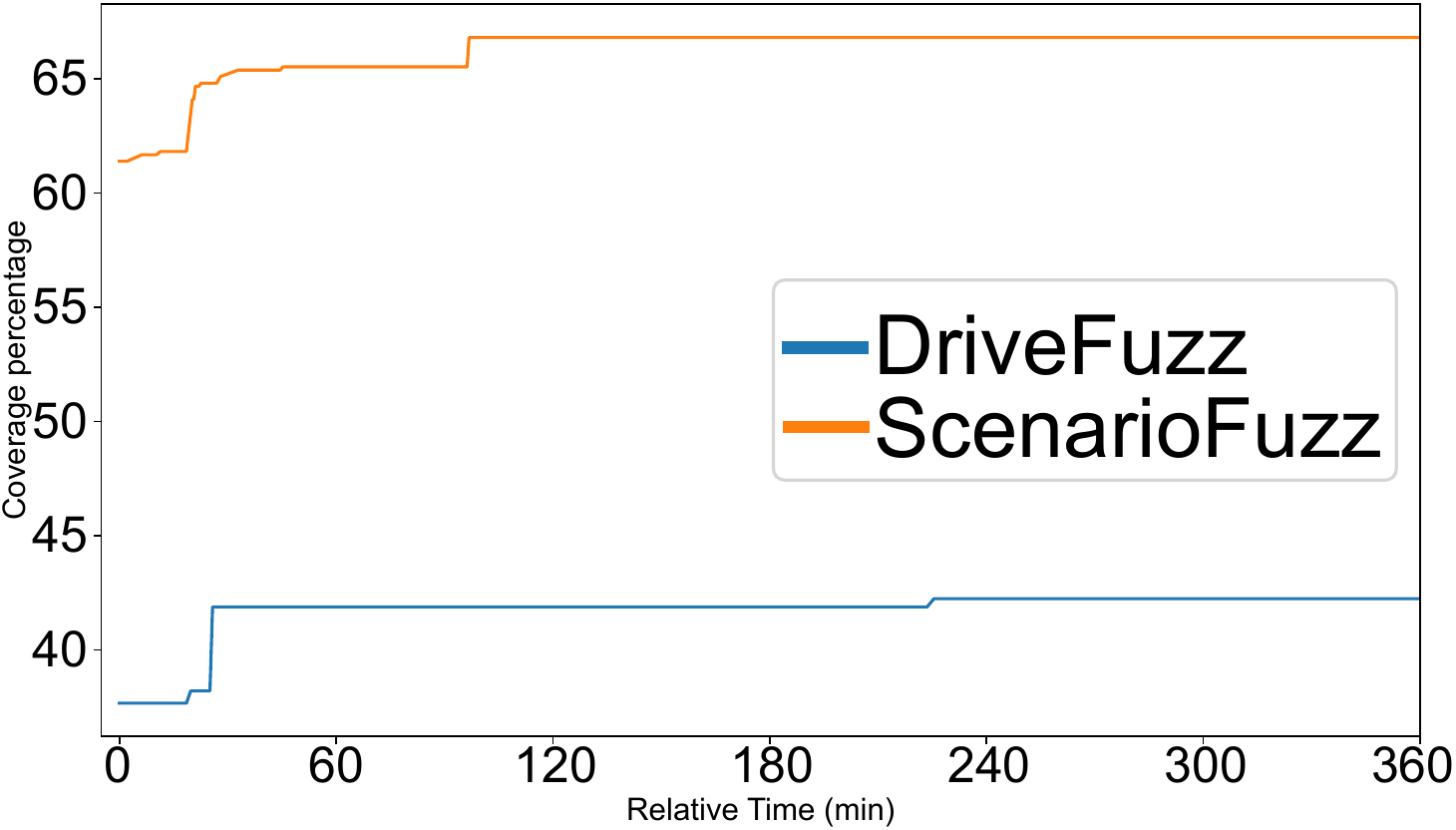}\label{fig10-2}}
	\hspace{0.08cm}  
	\subfloat[Behavior]{\includegraphics[width=0.15\textwidth]{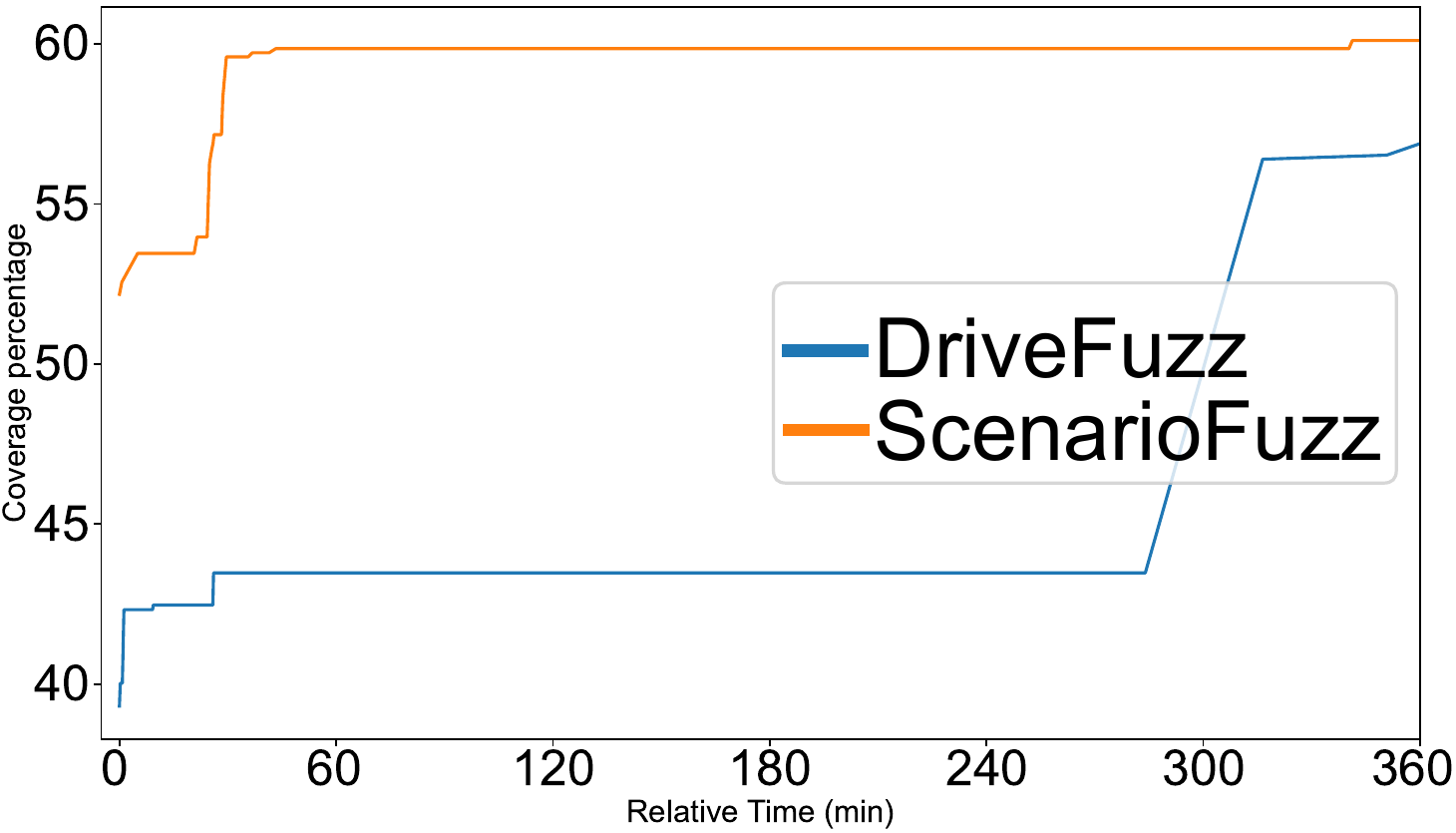}\label{fig10-3}}
	\vspace{-10pt}
	\caption{Comparison of code coverage among the two methods under different systems.}
	\label{fig10}
	\vspace{-10pt}
\end{figure}

\begin{figure}[h!]
	\centering
	\subfloat[Unreasonable placement]{\includegraphics[width=0.155\textwidth]{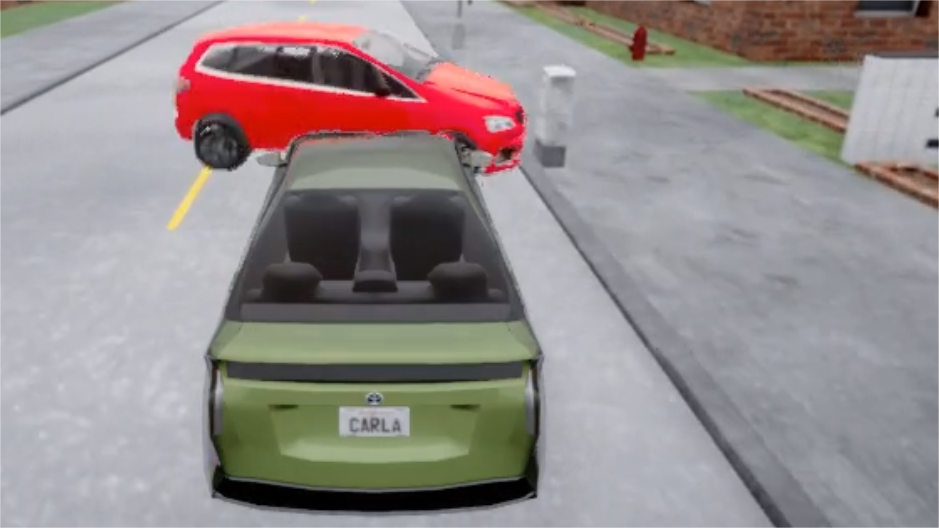}\label{fig11-1}}
	\hspace{0.01cm}  
	\subfloat[Incorrect placement]{\includegraphics[width=0.155\textwidth]{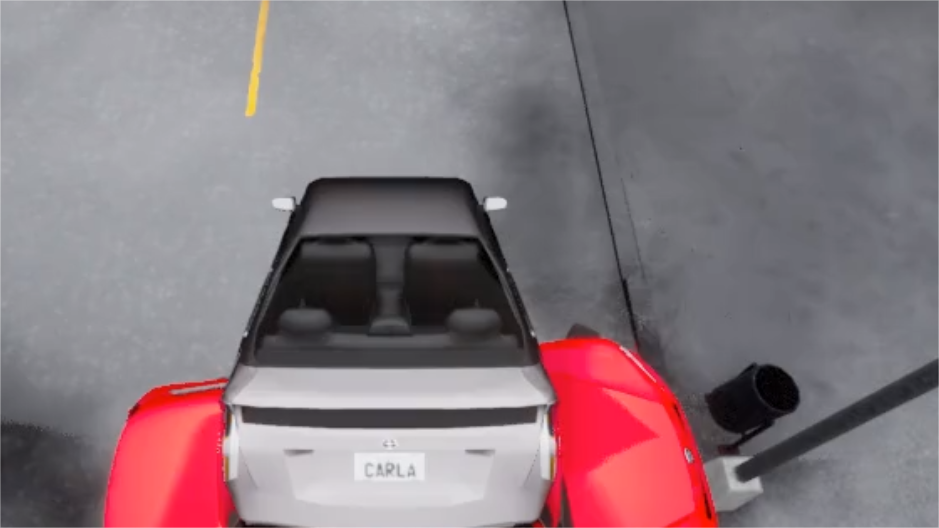}\label{fig11-2}}
	\hspace{0.01cm}
	\subfloat[Unreachable path]{\includegraphics[width=0.155\textwidth]{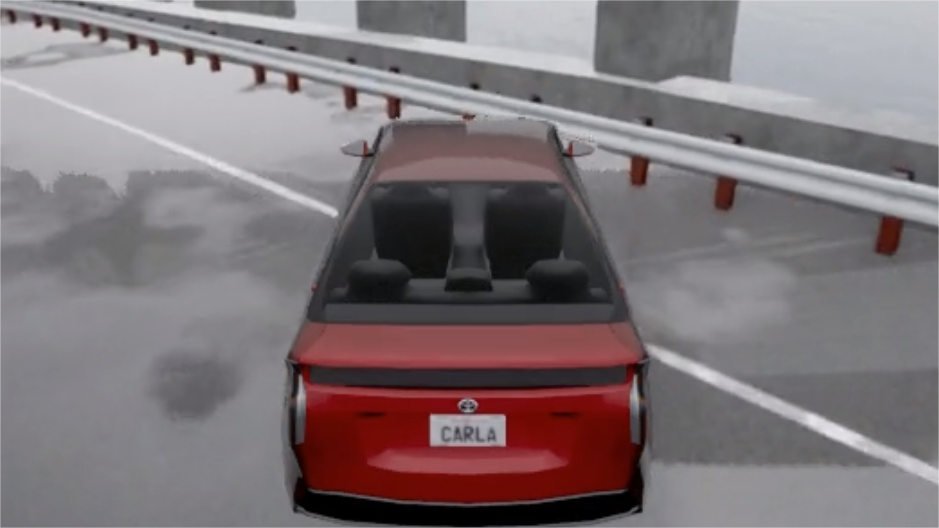}\label{fig11-3}}
	\vspace{-10pt}
	\caption{Wrong placement and path settings.}
	\label{fig11}
	\vspace{-15pt}
\end{figure}

As shown in Figure \ref{fig10}, the code coverage for Basic and Behavior was relatively low in the initial phase. This was due to scenarios generated by DriveFuzz, similar to those depicted in Figure \ref{fig11}, where ineffective scenarios reduced vehicle operation time, preventing the effective triggering of the corresponding code. During testing, DriveFuzz's entirely random mutation strategy led to it being stuck in local scenarios, continuously mutating on ineffective scenario seeds. However, as seen from the code line coverage results, our method triggers planning and control modules more frequently. This can be attributed to the corpus providing all appropriate placement points and path settings within scenarios, enhancing behavioral interactions, and thus enabling better triggering and coverage of related codes in these modules. Therefore, our method surpasses DriveFuzz in terms of code coverage rate.

\vspace{-15pt}

\textbf{Corpus}: The data from Figures \ref{fig8} and \ref{fig9}, along with Table \ref{tab1}, clearly demonstrate our corpus's pivotal role in augmenting Drivefuzz. Utilizing our corpus not only improves efficiency and the number of error scenarios generated but also reduces execution time. The corpus is fundamental to our work, ensuring rational positioning and movement paths for vehicles and pedestrians by extracting valid data from road networks, part of the HD map.

\begin{figure}[h!]
	\centering
	\subfloat[DriveFuzz]{\includegraphics[width=0.12\textwidth]{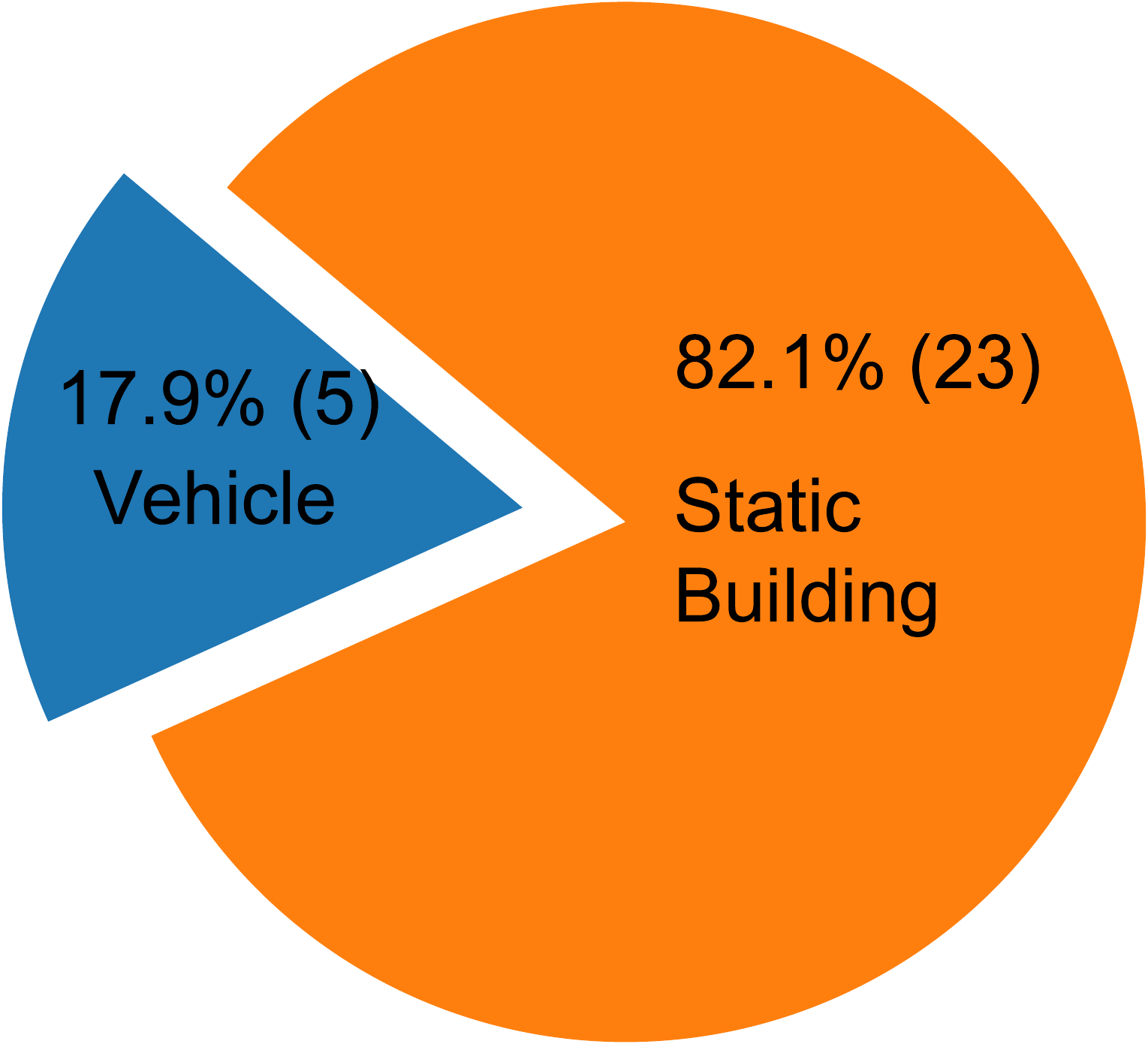}\label{fig12-1}}
	\hspace{0.08cm}  
	\subfloat[ScenarioFuzz]{\includegraphics[width=0.12\textwidth]{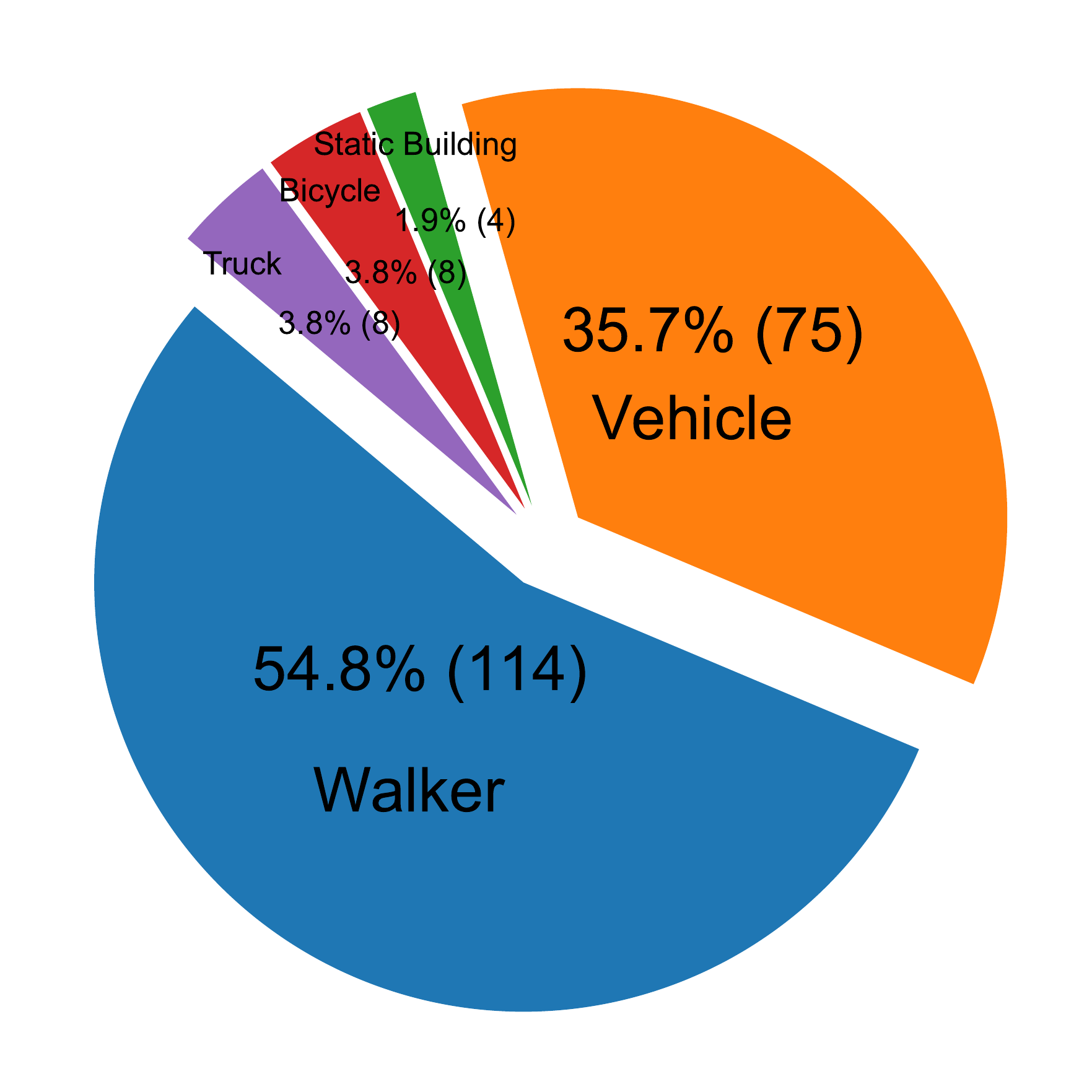}\label{fig12-2}}
	\vspace{-10pt}
	\caption{Distribution of collided object types generated by the two methods.}
	\label{fig12}
	\vspace{-15pt}
\end{figure}

\begin{figure*}[h!]
	\centering
	\includegraphics[width=0.9\textwidth]{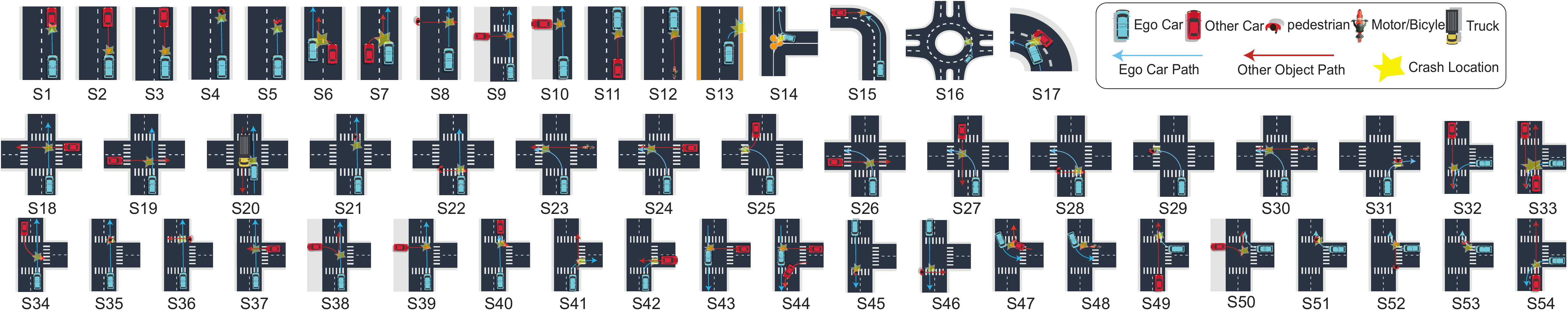} 
	\vspace{-10pt}
	\caption{Illustration of the error scenarios generated where collisions are prone to occur.}
	\label{fig13}
	\vspace{-10pt}
\end{figure*}

In contrast, Drivefuzz's methodology, which lacks such a corpus, randomly places vehicles and pedestrians within a threshold range of the ego car, with motion paths chosen randomly toward a destination. This approach presents several problems: 1) Random placements may lead to unreasonable or erroneous positions (Figures \ref{fig11-1}, \ref{fig11-2}), and 2) Random paths might be unreachable, causing navigation confusion and potential crashes (Figure \ref{fig11-3}). Moreover, our corpus enhances interactions, crucial for testing the ADS's perception and planning capabilities. An analysis of the collision types in scenarios generated by both methods (Figure \ref{fig12}) shows that Drivefuzz, lacking a robust information source, produces fewer varied collision scenarios, primarily with static structures.

\vspace{-10pt}

\vspace{-10pt}

\subsubsection{The Identified Bugs and Typical Scenarios}
\label{section5:2:5}

In our framework, we define a 'bug' as any deviation in the system's operation from its expected performance or safety standards. These deviations could manifest as reduced system functionality, potential safety hazards, or incorrect responses to environmental stimuli. We continuously monitored the operational status of various systems under generated error scenarios and identified 58 bugs, as shown in the supplementary material. This revealed serious safety issues in components across different modules of each system. We highlight a few concerning typical cases:

\textbf{Case 1}: Lidar failed to detect objects below a certain height, such as children and pedestrians lying on the ground (Autoware Bug2,4).

\textbf{Case 2}: Lidar object detection models (Autoware Bug11) and fusion positioning models (Autoware Bug21) were prone to errors due to lidar noise, such as reflections from walls, slopes, and trees, subsequently disrupting decision-making and control modules.

\textbf{Case 3}: In the CARLA, Autoware could not detect all traffic lights and signs, nor recognize certain types of pedestrians (Autoware Bug5-9). These deficiencies may stem from a lack of fine-tuning in the object detection model for simulated elements, highlighting the gap between simulation and reality.

\textbf{Case 4}: End-to-end autonomous driving systems showed vulnerabilities to certain colors, weather conditions, and lighting. For example, they failed to detect objects with red appearances (Transfuser Bug2), and the system's BEV visual prediction mechanism was severely impaired at night or when multiple objects were close together (NEAT Bug1). Additionally, there were limitations in translating neural network model outputs from predicted waypoints to control signals (LAV Bug7) and data loss in the Lidar-to-image conversion process, leading to stasis (Transfuser Bug1).

\textbf{Case 5}: All systems lacked flexible local planning and emergency response capabilities, with a poor understanding of traffic rules for lateral traffic at intersections. In such cases, Autoware assumed right of way, refusing to yield, while other systems adopted a conservative driving approach, opting to remain stationary, leading to collisions under some confrontational driving behaviors.

The discovery of these bugs was facilitated by our method's mutators, which adjusted for external weather and lighting conditions, as well as object shapes and colors. Additionally, identifying issues in the planning and control modules was made possible by the corpus, which provided reasonable placement and motion paths for the ego car and other objects. This encouraged diverse interaction patterns and ensured coverage of control signals (see supplementary materials). Faced with a multitude of generated error scenarios, particularly collision scenarios, we utilized the self-supervised trajectory clustering method introduced in \S\ref{section4:8}. This involved clustering scenario data and manually categorizing them, ultimately identifying 54 types of scenarios likely to lead to collisions, as shown in Figure \ref{fig13}. It is important to note that these 54 categories represent a high-level summary of all collision scenarios and do not indicate repetitiveness. The same interaction could occur in different contexts, such as: 1) at different locations; 2) under varying weather conditions; 3) with differently shaped or colored collision objects; 4) influenced by other objects; 5) with distinct trajectories leading to varied collision points. As shown in Table \ref{tab0}, our method demonstrated diversity in the data from experiments on a single seed. For example, in scenario S35, pedestrian collisions occurred, featuring pedestrians of different appearances and ages. In the supplementary material, we have illustrated the trajectory and error distribution when testing the Autoware system across various maps.

\vspace{-10pt}
\section{Related Work}

Current approaches in ADS testing often rely on predefined scenario sources, necessitating exact scenario layouts, starting and ending points for vehicles and pedestrians, and constraints on their behaviors. This method typically requires choreographing a starting scenario with designated constraints, which can be time-consu\-ming and limits the variety of scenarios to those pre-designed. To achieve comprehensive testing, multiple scenarios must be pre-set, but this approach is not exhaustive.

AV-Fuzzer was an early method based on this approach, utilizing a genetic algorithm to generate object trajectories within a starting scenario \cite{9251068}. SAMOTA adopts a manual approach, annotating road sections for scenario locations and testing through parameter combinations and search \cite{haq2022efficient}. In contrast, DriveFuzz departs from relying on a predefined scenario source. It utilizes CARLA API's placement waypoints for randomizing the starting and ending points of the ego car and nearby objects, driving seed mutation and generation through driving score feedback \cite{Kim22DriveFuzz}.

Our method, in response, employs map crawling technology to extract placement positions and reasonable driving paths based on road networks, proposing a novel corpus construction method. The mutator settings cover all scenario levels within CARLA's interface range, offering diversified object appearances in mutations. A two-stage mutation strategy effectively explores global and local areas, and with the addition of SEM, historical test data is used to enhance seed filtering, improving the overall efficiency of the method.

\vspace{-10pt}
\section{Conclusion}

In this paper, we introduced ScenarioFuzz, a fuzz testing generation framework based on autonomous driving scenarios. This framework aims to identify error scenarios in autonomous driving systems without an initial scenario source and reveal the underlying issues. Our work includes: (1) Automatically building a scenario seed corpus using map crawling technology, providing an effective information foundation for subsequent mutations; (2) Training a scenario evaluation model using historical test data, effectively filtering mutated scenario seeds, thereby enhancing the accuracy and efficiency of the fuzz testing generation process; (3) Designing a comprehensive fuzz testing framework to test autonomous driving systems from a scenario perspective. Our work can effectively mine error scenarios of ADS and locate a series of bugs.


\bibliographystyle{ACM-Reference-Format}
\bibliography{acmart.bib}

\end{document}